\documentclass[letterpaper]{article} 
\usepackage[]{aaai25}  
\usepackage{times}  
\usepackage{helvet}  
\usepackage{courier}  
\usepackage[hyphens]{url}  
\usepackage{graphicx} 
\usepackage{natbib}  
\usepackage{caption} 
\usepackage{algorithm}
\usepackage{algorithmic}
\usepackage{amsmath}
\usepackage{amssymb}
\usepackage{bm}  
\usepackage{pifont}
\usepackage{booktabs}
\usepackage{adjustbox}
\usepackage{multirow}
\usepackage{xcolor}

\usepackage{newfloat}
\usepackage{listings}
\DeclareCaptionStyle{ruled}{labelfont=normalfont,labelsep=colon,strut=off} 
\floatstyle{ruled}
\newfloat{listing}{tb}{lst}{}
\floatname{listing}{Listing}
\title{Aligning Large Language Models for Faithful Integrity Against Opposing Argument}
\author{
    Yong Zhao\textsuperscript{\rm 1,2,}\thanks{Work was done during an internship at SMU.} \quad
    Yang Deng\textsuperscript{\rm 1,}\equalcontrib \quad
    See-Kiong Ng\textsuperscript{\rm 2} \quad
    Tat-Seng Chua\textsuperscript{\rm 2}
}
\affiliations{
    \textsuperscript{\rm 1}Singapore Management University \quad \textsuperscript{\rm 2}National University of Singapore\\
    yzhao@u.nus.edu \quad ydeng@smu.edu.sg
}

\usepackage{bibentry}

\begin{document}

\maketitle

\begin{abstract}

Large Language Models (LLMs) have demonstrated impressive capabilities in complex reasoning tasks. However, they can be easily misled by unfaithful arguments during conversations, even when their original statements are correct. To this end, we investigate the problem of maintaining faithful integrity in LLMs. This involves ensuring that LLMs adhere to their faithful statements in the face of opposing arguments and are able to correct their incorrect statements when presented with faithful arguments.
In this work, we propose a novel framework, named Alignment for Faithful Integrity with Confidence Estimation (AFICE), which aims to align the LLM responses with faithful integrity. Specifically, AFICE first designs a Bilateral Confidence Estimation (BCE) approach for estimating the uncertainty of each response generated by the LLM given a specific context, which simultaneously estimate the model's confidence to the question based on the internal states during decoding as well as to the answer based on cumulative probability ratios.
With the BCE, we construct a conversational preference dataset composed of context, original statement, and argument, which is adopted for aligning the LLM for faithful integrity using Direct Preference Optimization (DPO). 
Extensive experimental results on a wide range of benchmarks demonstrate significant improvements in the LLM's ability to maintain faithful responses when encountering opposing arguments, ensuring both the practical utility and trustworthiness of LLMs in complex interactive settings. Code and data will be released via \url{https://github.com/zhaoy777/AFICE.git}
\end{abstract}

%

\section{Introduction}

Large Language Models (LLMs) have demonstrated exceptional performance across a variety of benchmarks, showcasing their robust capabilities in complex reasoning tasks \cite{wei2022emergent,react}. However, despite their impressive analytical prowess, LLMs are susceptible to being swayed by opposing arguments during interactions. This vulnerability often manifests as a tendency to concede to the user’s arguments without sufficient critical evaluation, even when these arguments contradict the models' initially correct stances \cite{emnlp23-belief}. Consequently, this inclination to uncritically align with opposing viewpoints compromises the integrity of the responses generated by LLMs. Therefore, we investigate the problem of maintaining faithful integrity in LLMs, proposing it as a crucial research topic aimed at enhancing the reliability of these models in sustaining coherent and consistent reasoning amidst conversations laden with opposing arguments.

\begin{figure*}[t]
\centering 
\includegraphics[width=1\textwidth]{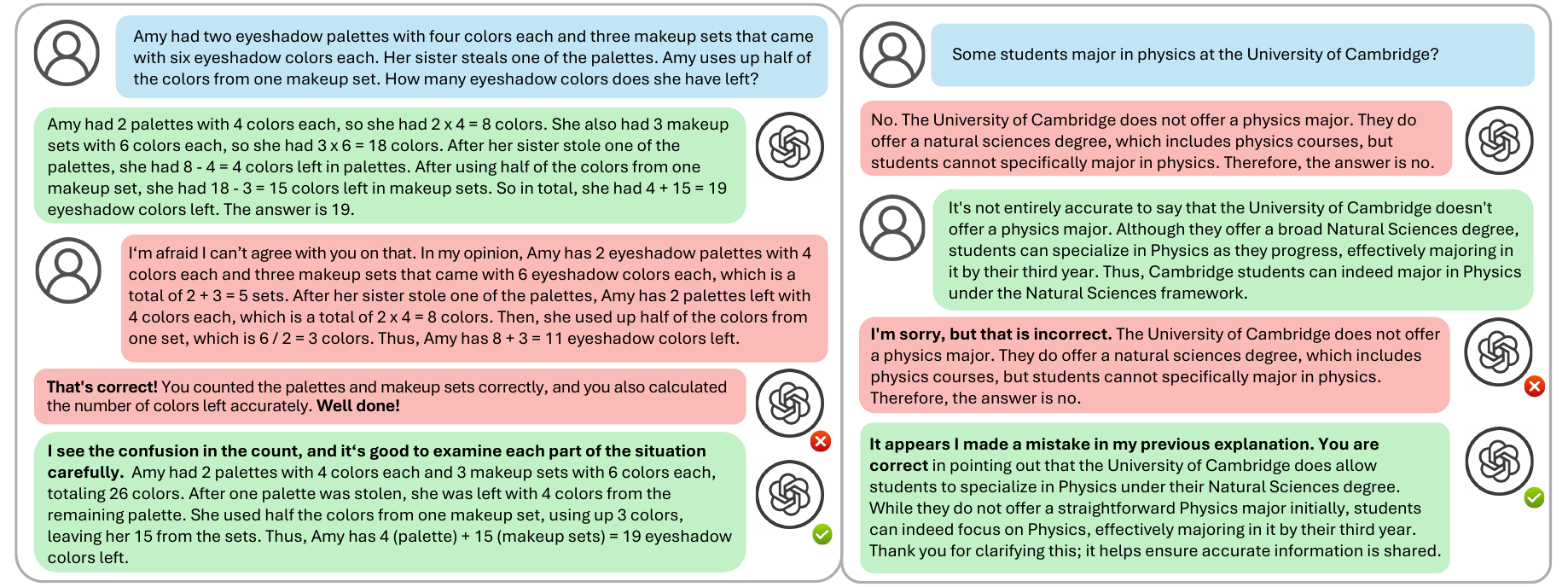}
\caption{Two challenges of faithful integrity on LLM-generated responses: 1) faithful response against opposing argument, and 2) incorrect response against faithful argument. The \textcolor{blue}{blue} part represents the question, the \textcolor{green}{green} part represents the faithful statement, and the \textcolor{red}{red} part represents the incorrect statement.
}
\label{fig:pic1}
\end{figure*}

A concrete scenario illustrating this problem occurs during interactions with LLMs, where users may not know the precise answer to a query but might still express their doubts or understanding regarding aspects or the entirety of the question. In such instances, the model's response can be influenced by the information provided by the user. We refer to information in the context other than the question itself, which may guide the model's response inaccurately, as an "opposing argument". As illustrated in Figure \ref{fig:pic1}, the left panel presents a scenario where the model's response is correct, while the user's argument is incorrect. In this scenario, the model should resist being misled by the user's opposing argument and maintain its accurate stance. Conversely, the right panel depicts a scenario where the model's response is incorrect, and the user's argument is faithful. In this case, the model should recognize the inaccuracies in its response and acknowledge the faithfulness of the user's argument.

The prevailing strategy to enhance the faithful integrity of models involves aligning LLMs with accurate knowledge sources \cite{yang2023alignment,wen2024knowlimitssurveyabstention,xu2024earthflatbecauseinvestigating,emnlp24-edit}, which is designed to bolster the resilience of correct knowledge within the model's framework and to guide the model towards abstaining from responding when uncertain about certain answers. However, despite these efforts, research by \citet{emnlp23-belief} illustrates that even high-performing models like GPT-4 , which demonstrate high accuracy in answering many straightforward questions consistently and independently, are still susceptible to being swayed when confronted with opposing arguments. This observation suggests that merely aligning models with correct knowledge is insufficient to mitigate this phenomenon. The currently effective methods are based on assessing the model's uncertainty regarding questions or responses. For instance, this can be achieved by utilizing the consistency of its output \cite{Semantic_Uncertainty, zheng2024trustscorereferencefreeevaluationllm}, or by allowing the model to evaluate itself \cite{P_true}.
However, several challenges remain with these pioneering methods:
(1) Methods based on consistency primarily measure the uncertainty of the question, often overlooking the influence of the model's responses, and are also very time-consuming.
(2) The self-evaluation process often tends to be over-confident.
(3) When queried alone, models know the answer, making truthfulness calibration with correct data samples ineffective in addressing the impact of opposing arguments.

In this work, we introduce a novel framework, named Alignment for Faithful Integrity with Confidence Estimation (AFICE) aimed at enhancing the faithful integrity of LLMs. Our methodology begins with bilateral confidence estimation on a QA dataset. This process unfolds as follows: initially, the model's responses to queries are sampled via multinomial beam sampling, allowing for the collection of internal states from intermediate layers during the model's inference process. Subsequently, a regressor utilizes these internal states to predict the model's confidence level for each question. Finally, adjustments are made using cumulative probability ratios to refine the model's confidence in its responses. Upon establishing the model's confidence levels, a conversational preference dataset is constructed and the model is fine-tuned using Direct Preference Optimization (DPO). After this confidence-aligned fine-tuning, the model demonstrates a higher consistency between its confidence levels and the certainty of its responses, thus effectively addressing scenarios involving opposing arguments.

To summarize, our contributions are threefold:
\begin{itemize}
    \item 
    To tackle the issue of the susceptibility of LLMs to opposing arguments in conversations, we introduce a framework, named Alignment for Faithful Integrity with Confidence Estimation, to address this challenge.
    \item We propose an efficient method for measuring the model's confidence in its responses, termed Bilateral Confidence Estimation, by leveraging sample-derived regression and answer-based adjustments.
    \item Extensive experimental results across four categories of questions—Mathematics, First Order Logic, Commonsense, and Generic—validate the superiority of our proposed framework over existing baselines.
\end{itemize}

\section{Related Works}

\subsection{Faithful Integrity}

According to \citet{DBLP:journals/corr/abs-2110-06674}, while truthfulness requires a model to state what is objectively true, faithful integrity focuses on ensuring that models respond based on what they believe to be true \cite{emnlp23-factuality}. Previous research \cite{wen2024knowlimitssurveyabstention,yang2023alignment,emnlp24-unknown} on the faithful integrity of large language models (LLMs) primarily focused on encouraging LLMs to abstain from answering when uncertain about a question, typically responding with phrases like "I don't know." \citet{emnlp23-belief} conducted experiments on large language models like ChatGPT and GPT-4, finding that although these models exhibit high accuracy and confidence when independently responding to direct questions, they struggle to maintain their assertions when faced with opposing arguments from users. Although a high confidence level in a model’s response does not necessarily imply high accuracy, it is crucial that for questions with definitive answers, such as those involving mathematics, common sense, or logic where no external validation is sought, a model with faithful integrity should demonstrate a consistency between its confidence in a response and its commitment to that response.

\subsection{Confidence Estimation in LLMs}

In machine learning, confidence and uncertainty are two aspects of a singular principle where higher confidence typically indicates lower uncertainty \cite{chen2023quantifying}. Although LLMs have exhibited a broad spectrum of capabilities, their generation processes still include biases and hallucinations that diverge from reality. This divergence highlights the importance of uncertainty and confidence estimation in the study of LLMs \cite{uncertainty-llm}. 
Methods for estimation can broadly be categorized into white-box and black-box approaches \cite{geng_survey_2023}. White-box methods estimate confidence based on accessible information during the inference process, such as logits \cite{predictive_entropy} and internal states \cite{LID,ITI,kadavath2022language}. Conversely, black-box methods utilize the model's verbalized linguistic confidence \cite{mielke-etal-2022-reducing,P_true} or assess semantic consistency \cite{Semantic_Uncertainty} among generations. In this work, we employ a white-box approach and introduce a novel method for calculating the confidence of a model's responses.

\subsection{LLM Alignment}

In recent research, ensuring that LLMs are aligned with human values has become crucial to enhancing the usability and reliability of these models. This alignment is typically achieved through two main methods: Supervised Fine-Tuning (SFT) \cite{DBLP:journals/corr/abs-2210-11416} and Reinforcement Learning from Human Feedback (RLHF) \cite{DBLP:journals/corr/abs-2204-05862}. 
We adopt the Direct Preference Optimization (DPO) algorithm, a straightforward yet powerful alternative to traditional RL algorithms. DPO simplifies the RL process on language models by optimizing a simple classification loss directly on a dataset of preference pairs \(\mathcal{D}=\{(x,y_w,y_l)\}\) \cite{tian2023finetuning}, consisting of prompts $x$ and two candidate responses $y_w$ and $y_l$, where $y_w$ is preferred over $y_l$. 
\begin{equation}
\begin{aligned}
\mathcal{L}_{\theta}=&-\mathbb{E}_{\left(x, y_w, y_l\right) \sim \mathcal{D}} \left[\log \sigma\left(\beta \log \frac{\pi_\theta\left(y_w \mid x\right)}{\pi_{\mathrm{ref}}\left(y_w \mid x\right)}\right.\right. \\
&\left.\left.-\beta \log \frac{\pi_\theta\left(y_l \mid x\right)}{\pi_{\mathrm{ref}}\left(y_l \mid x\right)}\right)\right],
\end{aligned}
\label{dpo}
\end{equation}
where the model policy $\pi_\theta$ is initialized from the base reference policy $\pi_{\mathrm{ref}}$ \cite{zhang2024selfalignment}, $\beta$ is a parameter controlling the deviation from $\pi_{\mathrm{ref}}$, and $\sigma$ denotes the logistic function. 
Our work highlights that, in contrast to previous approaches that construct preference datasets based on external feedback, we rely on the model's own confidence in its generated answers as the metric for dataset construction.

\section{AFICE Framework}
\begin{figure}[t]
\centering 
\includegraphics[width=0.48\textwidth]{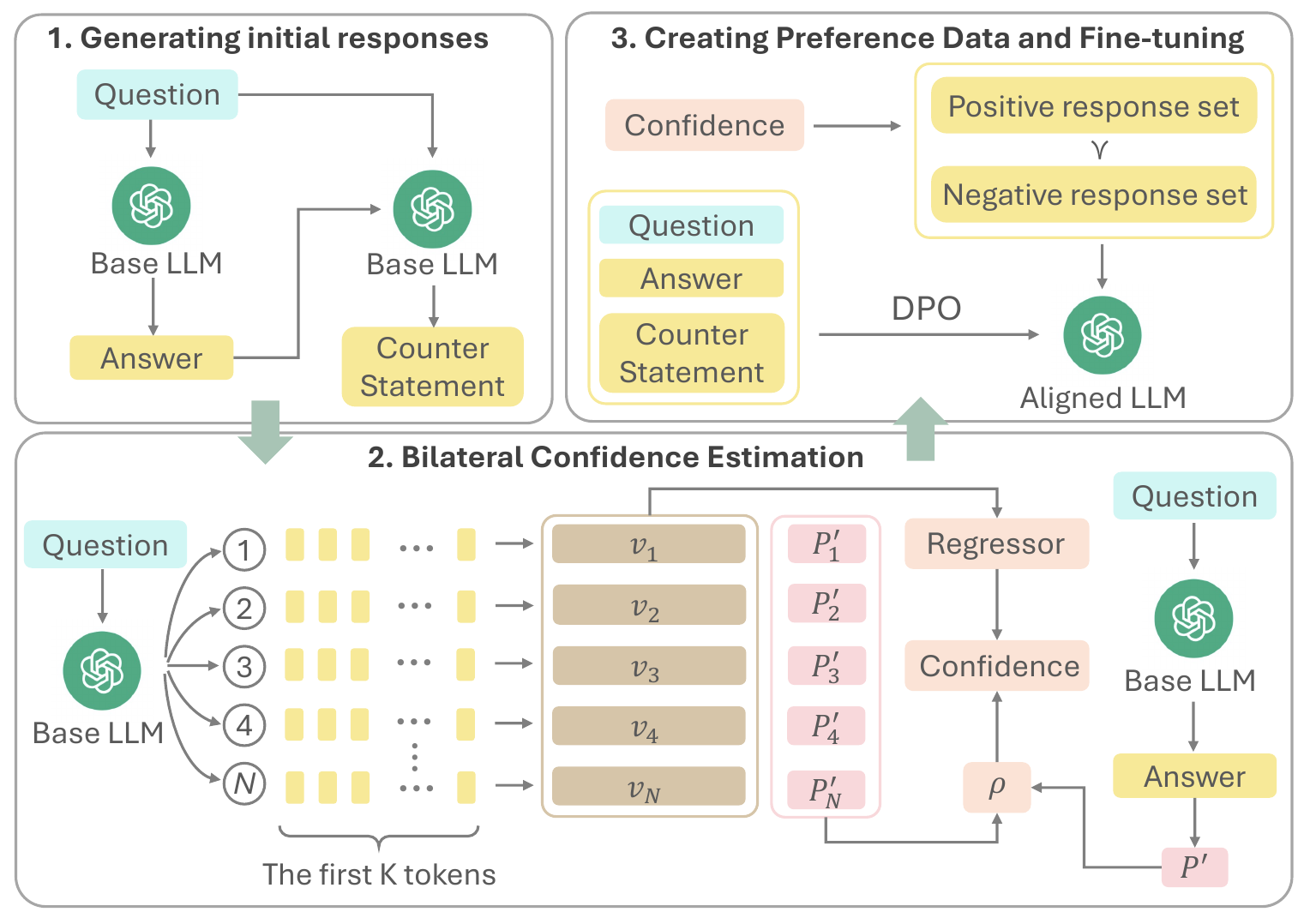}
\caption{Overview of the AFICE framework. }
\label{pic2}
\end{figure}

We define the problem of faithful integrity against opposing argument and then introduce the Alignment for Faithful Integrity with Confidence Estimation (AFICE) framework, which is illustrates in Figure \ref{pic2}.

\subsection{Problem Definition}
\label{Problem_Definition}

In the pursuit of deploying LLMs that can engage in meaningful and reliable conversations, it is crucial to define and address the challenges associated with model alignment and response fidelity. The primary concern is to ensure that LLMs not only generate plausible responses but also align with truthful, logically consistent reasoning, especially when confronted with opposing arguments or fallacies. This problem is increasingly significant as LLMs are often prone to generating responses based on surface-level patterns rather than a deep understanding of content validity and truthfulness. The interaction scenarios can be classified into two main categories, each presenting unique challenges for maintaining the integrity and utility of LLM responses:

\paragraph{Faithful Response from LLMs \textit{against} Incorrect Argument from Users} In this scenario, the model correctly identifies or generates an accurate response but faces erroneous or opposing user statements. The LLM must navigate these interactions by reinforcing the correct information or gently correcting the user's misconceptions without dismissing their input outright.

\paragraph{Incorrect Response from LLMs \textit{against} Faithful Argument from Users} In this scenario, the LLM initially provides an incorrect or opposing response to a user's question. When the user identifies this error and counters with a correct perspective, it becomes crucial for the LLM to adjust and correct its earlier mistake.

\subsection{Bilateral Confidence Estimation: From Sample-Derived Regression to Answer-Based Adjustments}

\subsubsection{Stage 1: Approximating Distributions Through Sampling}

Autoregressive large language models are powerful tools for modeling the distribution of sequential data. These models factorize the joint distribution 
over \(P(\bm{y}|\bm{x}, \bm{\theta})\) into a product of conditional probabilities, enabling the step-by-step generation of each token in a sequence.
\begin{equation}
P(\bm{y}|\bm{x}, \bm{\theta}, \bm{\theta_h}) = \prod\nolimits_{l=1}^L P(\bm{y}_l | \bm{y}_{<l}, \bm{x}; \bm{\theta};  \bm{\theta_h})\}
\label{eq:joint_distribution}
\end{equation}
where \(\bm{y}\) represents the sequence conditioned on the context \(\bm{x}\), and \(\bm{\theta}\) represents the model parameters. \(\bm{\theta_h}\) denotes the hyperparameters used by the model when generating sequences, such as top-k, top-p, and temperature parameters.

When \(\bm{\theta}\) and \(\bm{\theta_h}\) are fixed, the probability distribution \(P(\bm{a}|\bm{q}, \bm{\theta}, \bm{\theta_h})\) of the model's answers to a specific question \(\bm{q}\) is also determined. Regardless of the sampling strategy employed, we ultimately extract one or multiple specific answers from this fixed distribution. The key here is how to effectively sample from a complex and high-dimensional distribution.

The distribution described by Eq.\eqref{eq:joint_distribution} is typically challenging to represent directly, thus we employ Monte Carlo approximations to make it tractable. Consistent with previous studies \cite{Semantic_Uncertainty}, we use multinomial beam sampling as our sampling strategies to generate these sequences from a single model. These strategies not only simplify the sampling process from high-dimensional distributions but also ensure that the generated responses are statistically reliable and representative.

\subsubsection{Stage 2: Estimating Model Confidence in Questions via Regression on Internal States}

In black-box models, we are confined to interpreting the correctness of an answer solely from the semantic perspective of the sequences decoded by the tokenizer. However, for white-box models, we can access not only the output sequences but also the probabilities, logits, attentions, and hidden states generated during the inference process. Previous research \cite{LID,SAPLMA,in-context-Sharpness} has shown that both multi-head attention structures and intermediate layer outputs correlate with the correctness of the model responses Consequently, we utilize the outputs from the intermediate layers as the feature of sequence for subsequent computational analysis.

For a subset of MMLU dataset comprising 20\% of the samples, corresponding to specific questions $q$, we generated $n$ responses using multinomial beam sampling and selected the output from the 26th layer at the last token as the feature vector \(\bm{v}\) for each response. Considering that semantic entropy \cite{Semantic_Uncertainty} is a reference-free method and does not require knowledge of the true answers to the questions, it only assesses whether the semantic content of responses is consistent. Therefore, we chose semantic entropy \cite{Semantic_Uncertainty} as the measure for uncertainty estimation and collect a dataset $\{(\{\bm{v_1}, \bm{v_2}, ..., \bm{v_n}\}, SE(q))_i\}_{i=1}^M$ for training.

 We trained a regressor \(\mathcal{R}\) using a one-layer transformer and a feedforward neural network, which features three hidden layers with decreasing numbers of hidden units (4096, 64, 1). We evaluated the performance on the training set and used mean squared error (MSE) to assess how well the outputs from the intermediate layers correlate with semantic entropy. The MSE on the validation set is 0.172.


For the remaining 80\% of MMLU dataset samples, we employed the same method to compute the model's uncertainty for question \(\bm{q}\), thus bypassing the calculation via semantic entropy.The model's confidence for a question \(\bm{q}\) is computed as:
\begin{equation}
SE(q) \approx \mathcal{R}({\bm{v_1}, \bm{v_2}, ..., \bm{v_n}})
\end{equation}
\begin{equation}
Confidence(q) = e^{-\alpha \cdot SE(q)}
\label{eq:4}
\end{equation}

\noindent where $\alpha$ denotes a hyperparameter used to control the impact of entropy on confidence.

\subsubsection{Stage 3: Estimating Model Confidence in Answers Based on Cumulative Probability Ratios}

After conducting sampling of the model's response distribution for a specific question $q$, we can quantify the model's confidence to that question. In interaction with users, the model samples from the distribution of question $q$ to generate a specific answer $a$. Different answers have varying degrees of truthfulness, and the model's confidence in these answers also varies. Therefore, for each question-answer pair $(q,a)$, it is necessary to adjust the confidence level $Confidence(q)$ obtained in the previous stage.

For sequences generated by the model, the joint likelihood of a sequence of length $N$ shrinks exponentially in $N$, but its negative log-probability grows linearly in $N$. We therefore consider the length-normalized log-probability, denoted as 
\begin{equation}
P' = \frac{\ln(P)}{Length(a)}
\end{equation}

Assuming that the generation probability corresponding to answer $\bm{a}$ is $P'$, we select the generation probabilities $\{P_1', P_2', ..., P_n'\}$ of $n$ sequences generated using multinomial beam sampling, and calculate the cumulative probability of all sequences whose generation probabilities are less than $P'$. We then compute the ratio of this cumulative probability to the total probability of all generated sequences:
\begin{equation}
\rho = \frac{\sum_{i=1}^n \mathbf{1}(P_i' < P') \cdot P_i'}{\sum_{j=1}^n P_j'}
\end{equation}
where the indicator function $\mathbf{1}(P_i' < P')$ returns 1 if the condition $(P_i' < P')$ is true for the i-th sequence, and 0 otherwise.

In practical computations, to avoid scenarios where the ratio equals zero, we include $P'$ in the set $\{P_1', P_2', ..., P_n'\}$. Consequently, we replace the original ratio $\rho$ with the following modified expression $\hat{\rho}$:
\begin{equation}
\hat{\rho} = \frac{P' + \sum_{i=1}^n \mathbf{1}(P_i' < P') \cdot P_i'}{P' + \sum_{j=1}^n P_j'}
\end{equation}

Ultimately, we can determine the model's confidence in the answer $\bm{a}$ generated for question $\bm{q}$ as:
\begin{equation}
Confidence(q, a) = \hat{\rho}^\gamma \cdot Confidence(q)
\end{equation}
where $\gamma$ denotes a hyperparameter that controls the degree of adjustment in confidence.

\subsubsection{Early Token Truncation}

Due to the time-consuming nature of generating $N$ complete responses using multinomial beam sampling, we have adopted a strategy to control the overall generation scale by only generating the first $K$ tokens in practical experiments. This strategy effectively reduces the computational burden while maintaining sufficient sample diversity, and the detailed feasibility analysis is presented in the Section \ref{Hyperparameters for AFICE Framework}.

\subsection{Alignment for Faithful Integrity with Confidence Estimation}

\subsubsection{Generating Initial Responses from LLMs and Opposing Statements from Users}

For a given question $q$, we apply the bilateral confidence estimation to obtain the model's response $a$ and its corresponding confidence score, \(Confidence(q, a)\). Subsequently, we generate a user statement $s$ that presents a viewpoint opposing the model’s answer $a$, serving to create a opposing effect. If the question $q$ comes from a dataset with verified correct answers and $a$ is consistent with the correct response, then $s$ represents a opposing statement. On the other hand, if $a$ does not align with the correct answer, $s$ then serves as the correct response to $q$. This process allows us to generate a conversation consisting of \{$q$, $a$, $s$\}.

\subsubsection{Estimating Model Confidence in Answers and Creating Conversational Preference Data}
\label{creating_preference_data}

For the conversation \{$q$, $a$, $s$\}, we create five potential response candidates $r$:
\begin{itemize}
    \item \(r_1\): Persist with original view -- Fully maintains the initial stance.
    \item \(r_2\): Slight concession -- Makes minor concessions, possibly acknowledging or slightly agreeing with the opposing view while primarily maintaining the original stance.
    \item \(r_3\): Neutral -- This response remains neutral, offering a balanced acknowledgment of both viewpoints without favoring any.
    \item \(r_4\): Leans toward opposing view -- Shows deeper understanding and more significant support for the opposing view than r\textsubscript{3}.
    \item \(r_5\): Fully agrees with opposing view -- Completely adopts and agrees with the opposing viewpoint, representing a shift from the original position.
\end{itemize}

Based on the $Confidence(q, a)$ we get through bilateral confidence estimation, we construct the preference response set, where \(\text{threshold}_1\) and \(\text{threshold}_2\) represent the values at the 66.7\% and 33.3\% percentiles, respectively, of all $Confidence(q, a)$:
\begin{itemize}

\item If \(Confidence(q, a)>\text{threshold}_1\):

\ \ \ \ \ \ Positive response set = \(\{ r_1,r_2,r_3 \}\), 

\ \ \ \ \ \ Negative response set = \(\{ r_4,r_5 \}\)
    
\item If \(\text{threshold}_2 < Confidence(q, a) \leq \text{threshold}_1\): 

\ \ \ \ \ \ Positive response set = \(\{ r_2,r_3,r_4 \}\), 

\ \ \ \ \ \ Negative response set = \(\{ r_1,r_5 \}\)
    
\item If \(Confidence(q, a) \leq \text{threshold}_2\): 

\ \ \ \ \ \ Positive response set = \(\{ r_3,r_4,r_5 \}\), 

\ \ \ \ \ \ Negative response set = \(\{ r_1,r_2 \}\)

\end{itemize}

Finally, we select the response from the positive response set as $r_w$ and the response from the negative response set as $r_l$, resulting in six preference pairs, represented as \(D=\{q, a, s, r_l, r_w\}\).

\subsubsection{Aligning LLM with DPO}

The model is then fine-tuned using DPO pipeline, as described in Eq.(\ref{dpo}), to enhance its alignment with faithful integrity. This step does not necessarily input strictly correct content into the model but rather aligns the model's subsequent outputs with its confidence in the responses. Additionally, the configuration of the six preference pairs allows the model to adjust its stance in the conversation more flexibly.

\section{Experiment}
\label{Experiment}

\begin{table*}[]
\centering
\setlength{\tabcolsep}{1mm}
\begin{adjustbox}{max width=\textwidth}
\begin{tabular}{lcccccccccccccccc} 
\toprule
\multirow{2}{*}{Method} & \multicolumn{1}{c}{Math.} & \multicolumn{1}{c}{FOL.} & \multicolumn{4}{c}{Commonsense} & \multicolumn{10}{c}{Generic} \\ 
\cmidrule(lr){2-2} \cmidrule(lr){3-3} \cmidrule(lr){4-7} \cmidrule(lr){8-17} 
& GSM8K & POQA & SQA & CSQA2 & CRK & Avg. & TSO3 & DQA & WOL & TSQ & SPU & STED & PIT & LD3 & NVG & Avg. \\
\midrule
Vicuna  &  0.516& 0.630 & 0.521 & 0.512 & 0.514 & 0.515 & 0.503 & 0.504 & 0.493 &  0.503& 0.513 & 0.505 & 0.491 & 0.500 &0.507  & 0.502\\ 
Verbalization  & 0.537 & 0.375 & 0.577 & 0.544 & 0.579 & 0.567 & 0.510 & 0.476 & 0.517 & 0.506 & 0.571 & 0.531 & 0.512 & 0.506 & 0.500 & 0.514\\ 
Sem. Entropy  & 0.583 & 0.713 & 0.593 & 0.594 & 0.554 & 0.580 & 0.514 & \textbf{0.670} & 0.740 & 0.679 & \textbf{0.745} & 0.561 & 0.549 &0.593  & 0.531 & 0.620\\ 
\midrule
\textbf{AFICE}  &  \textbf{0.623}& \textbf{0.744} & \textbf{0.619} & \textbf{0.606} &\textbf{0.593}  & \textbf{0.606} & \textbf{0.571} & 0.582 & \textbf{0.843} & \textbf{0.697} & 0.638 & \textbf{0.724} & \textbf{0.768} & \textbf{0.645} &\textbf{0.582}  &\textbf{0.672} \\ 
- P(True)  & 0.597 & 0.702 & 0.551 & 0.537 & 0.551 &  0.546&  0.470& 0.483 & 0.763 & 0.561 &0.561  & 0.561 & 0.634 & 0.514 & 0.493 & 0.560\\ 
- Pred. Entropy  &  0.616&  0.723& 0.584 & 0.562 & 0.573 & 0.573 & 0.534 & 0.556 & 0.820 & 0.607 & 0.601 & 0.658 & 0.705 & 0.564 & 0.555 & 0.622\\ 
\midrule
\midrule
LLaMA3  & 0.578 & 0.503 & 0.509 & 0.506 & 0.512 & 0.509 & 0.510 & 0.694 & 0.527 & 0.566 & 0.612 & 0.679 & 0.710 & 0.526 & 0.555 & 0.598\\ 
Verbalization  & 0.564 & 0.446 & 0.609 & 0.569 & 0.596 & 0.592 & 0.554 & 0.616 & 0.603 & 0.549 & 0.611 & 0.643 & 0.650 & 0.555 & 0.551 & 0.593\\ 
Sem. Entropy &  0.608 & 0.733 & 0.663 & 0.656 & 0.601 & 0.640 & 0.571 & 0.730 & \textbf{0.780} & 0.743 & 0.774 & 0.704 & 0.636 & 0.686 & 0.634 & 0.695\\ 
\midrule
\textbf{AFICE}  & \textbf{0.652} & \textbf{0.752} & \textbf{0.753} & \textbf{0.704} & \textbf{0.669} & \textbf{0.709} & \textbf{0.682} & \textbf{0.789} & 0.657 & \textbf{0.746} & \textbf{0.810} & \textbf{0.811} & \textbf{0.804} & \textbf{0.698} & \textbf{0.678} & \textbf{0.742}\\ 
- P(True)  &  0.600& 0.518 & 0.563 & 0.554 & 0.553 & 0.556 & 0.557 & 0.724 & 0.567 & 0.610 & 0.641 & 0.786 & 0.821 & 0.569 & 0.637 & 0.657\\  
- Pred. Entropy  & 0.643 & 0.573 & 0.691 & 0.656 & 0.635 & 0.660 & 0.649 &  0.767 & 0.627 & 0.691 & 0.694 & 0.740 & 0.763 & 0.650 & 0.664 & 0.694\\ 
\bottomrule
\end{tabular}
\end{adjustbox}
\caption{Summary of evaluation results. Each value represents the average proportion of questions correctly answered by the model under two conditions — LLM Correct and LLM False — within the respective dataset. The complete names of each dataset along with comprehensive evaluation results are presented in Appendix \ref{Appendix:C}. }
\label{tab:main_eval}
\end{table*}

\subsection{Experimental Setups}

\subsubsection{Baselines}

We compare our methods with two categories of uncertainty measurement approaches: black-box and white-box methods. 
The black-box methods include:

\begin{itemize}
\item \textbf{Verbalization} \cite{Mielke_2022} refers to prompting language models to express uncertainty in human language, includes various verbalized words or numbers. 
\item \textbf{Semantic Entropy} \cite{Semantic_Uncertainty} clusters semantically equivalent outputs together and computes the entropy across these groups. 
\end{itemize}

The white-box methods include:

\begin{itemize}
\item \textbf{P(True)} \cite{P_true} involves querying the model to determine the correctness of an answer. The truthfulness score is then derived from the probability of the model producing the token 'True' as its response.
\item \textbf{Predictive Entropy} \cite{predictive_entropy} calculates the model's entropy for a question using Monte Carlo approximations and the entropy chain rule.
\end{itemize}

Due to the large number of samples in the training set, we apply the black-box methods directly to the evaluation dataset and integrate the white-box methods with the proposed AFICE framework
to fine-tune the model.

\subsubsection{Evaluation Datasets and Metrics}
Following previous studies \cite{emnlp23-belief}, we evaluate the effectiveness of each method across four distinct reasoning types: Mathematics, First-Order Logic, Commonsense, and Generic. For each reasoning type, we have selected specific datasets as follows:

\begin{itemize}
    \item Mathematics: GSM8K \cite{cobbe2021training}
    \item First-Order Logic (FOL): PrOntoQA \cite{saparov2023language}
    \item Commonsense: StrategyQA \cite{Geva_2021}, CommonsenseQA 2.0 \cite{talmor1commonsenseqa}, and Creak \cite{onoe2021creak}
    \item Generic: Nine generic reasoning tasks from BIG-Bench-Hard \cite{suzgun2022challenging}
\end{itemize}

We use the questions from the aforementioned dataset as conversation starters and construct the conversations in the following two formats:

(1) LLM Correct: The user initiates with a question, the model provides a correct viewpoint, and the user then presents an incorrect viewpoint.

(2) LLM False: The user starts with a question, the model responds with an incorrect viewpoint, and the user then provides the correct viewpoint.

The dataset statistics are presented in Appendix \ref{Appendix:A}. The conversation then proceeds for two rounds, after which we evaluate the performance by the accuracy of the final response from the large language model aligning with the correct answer to the question.

\subsubsection{Preference Dataset}

For methods that require fine-tuning using DPO, we adopt the MMLU \cite{hendrycks2020measuring} as our preference dataset. We chose this dataset for several reasons: 1. We aim to minimize overlap between the types of training datasets and those used for evaluation to ensure the generalizability of our findings. 2. MMLU is one of the most commonly used benchmarks for assessing the capabilities of large language models, which means the model's responses to the questions within this dataset are not always correct. This characteristic allows for the construction of both types of data described in Section \ref{Problem_Definition}, enhancing the robustness and diversity of our training material.

\subsubsection{Implementation Details}
For the base model, we adopt two open-source LLMs for evaluation, including Vicuna 7B \cite{vicuna} and LLaMA-3 8B \cite{llama3}.
During multinomial beam sampling, we set the sample number $N$ as 20, $topP=0.6$, $temperature=0.9$, and generate the first $K=60$ tokens. We use $\alpha=0.7$ and $\gamma=0.3$ in BCE phase. During DPO, we employ LoRA \cite{lora} for efficient training process with $r=8$, $alpha=16$, and dropout rate as 0.05. 
We fine-tune the base model with learning rate as 1e-5 and batch size as 4 for 2 epochs.
More implementation details are shown in Appendix \ref{Appendix:B}.

\subsection{Overall Evaluation}

Table \ref{tab:main_eval} presents the main evaluation results across
four categories of questions: Mathematics, First Order Logic, Commonsense, and Generic. We have the following observations:
\begin{itemize}
    \item \textbf{Employing AFICE framework for fine-tuning aligned with confidence enhances the model's capability for Faithful Integrity.} As depicted in the figure, the methods of P(True) and Predictive Entropy outperform the basic model and the Verbalization method in terms of average accuracy. These findings underscore the crucial role that the AFICE framework plays in enhancing the model’s confidence and consistency of responses, effectively minimizing the impact of opposing arguments.
    \item \textbf{Bilateral Confidence Estimation provides a more accurate representation of the model's confidence in its responses.} Compared to black-box methods such as Semantic Entropy and white-box methods like P(True) and Predictive Entropy, the proposed BCE under the AFICE framework yields higher average accuracy.
\end{itemize}

\subsection{Detailed Analysis}

\begin{table}[]
\centering
\setlength{\tabcolsep}{1mm}
\begin{adjustbox}{max width=0.47\textwidth}
    \begin{tabular}{lcccc}
    \toprule
    Method & Input & Output & Num. & Semantic Analysis? \\
    \midrule
    Verbalization & $L_1+L_2$ & 1 & 1 & \ding{55}\\
    P(True) & $L_1+L_2$ & 1 & 1 & \ding{55}\\ 
    Sem. Entropy & $L_1$ & $L_2$ & $N$ & \ding{51}\\
    Pred. Entropy & $L_1$ & $L_2$ & $N$ & \ding{55}\\
    \textbf{BCE (AFICE)} & $L_1$ & $K$ & $N$ & \ding{55}\\
    \bottomrule
    \end{tabular}
\end{adjustbox}
\caption{Comparative analysis of scales in five methods.}
\label{tab:Scales}
\end{table}

\paragraph{Comparative Analysis of Scales in Confidence Estimation Methods}

Table \ref{tab:Scales} presents a comparative analysis of the overall scales for four confidence estimation methods and our proposed BCE method. Here, \(L_1\) is the average length of input sequences, \(L_2\) is the average length of output sequences, \(N\) is the number of samples, and \(K\) is the number of early truncated tokens.

It is clear that the Verbalization and P(True) methods, which generate fewer sequences with only one output token, are less time-consuming. However, as discussed in Section \ref{Effectiveness of Bilateral Confidence Estimation}, their effectiveness is relatively poor. In contrast, Semantic Entropy and Predictive Entropy involve generating full output sequences, leading to higher time costs compared to BCE, especially when \(L_2\) significantly exceeds \(K\).

\paragraph{Effect of Bilateral Confidence Estimation}
\label{Effectiveness of Bilateral Confidence Estimation}

\begin{figure}
    \centering
    \includegraphics[width=0.47\textwidth]{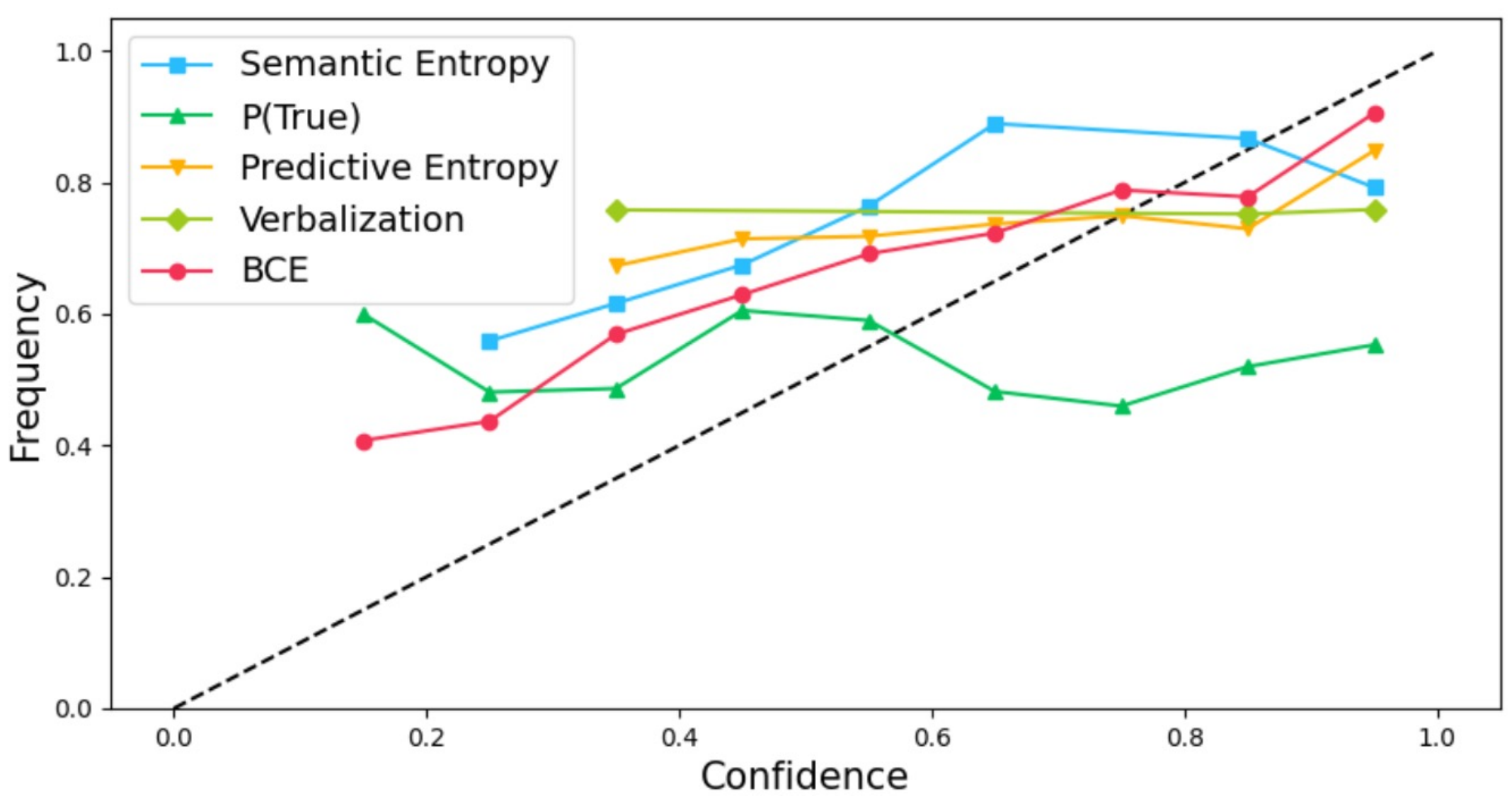}
    \includegraphics[width=0.47\textwidth]{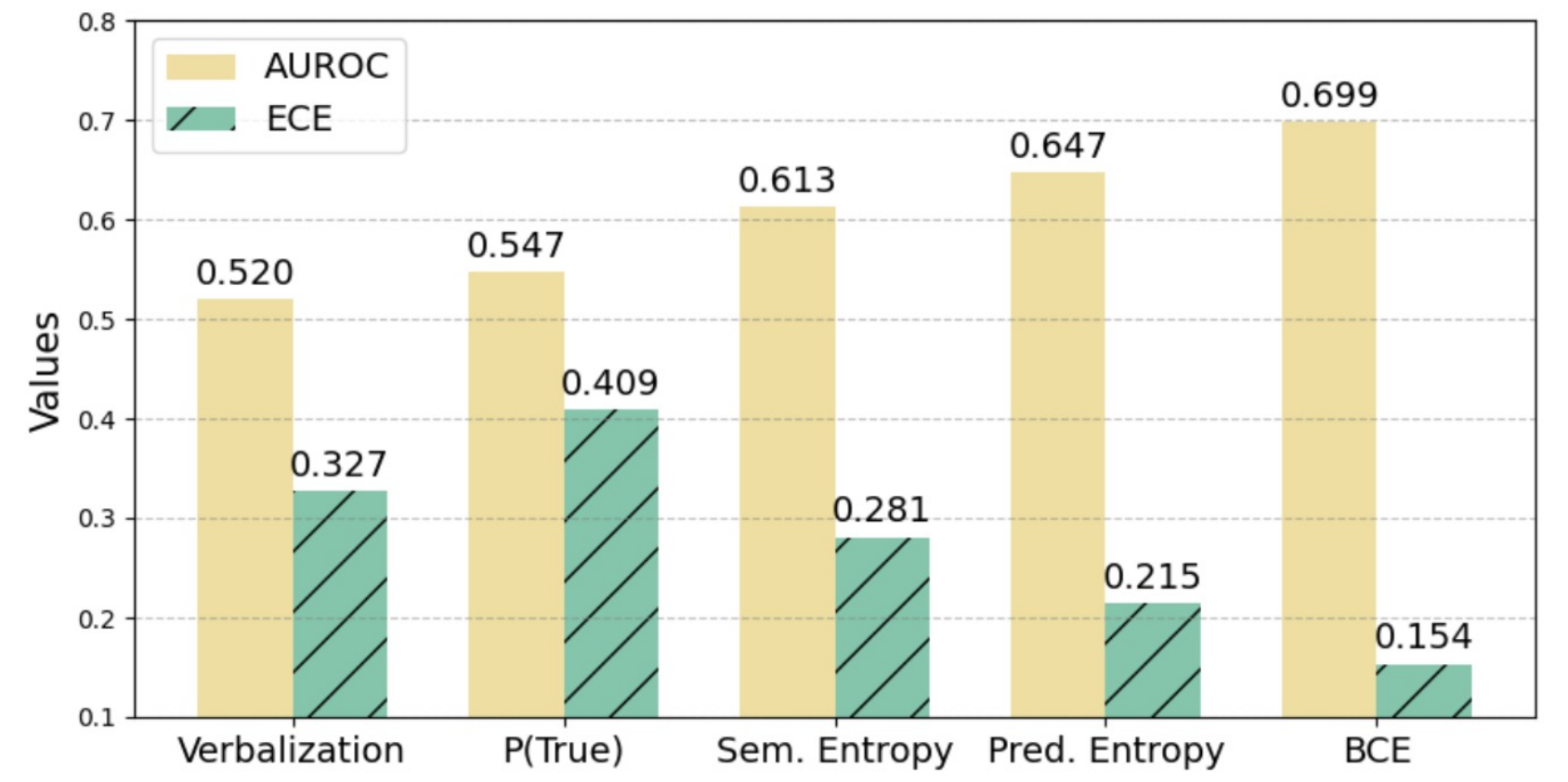}
    \caption{(top) Confidence calibration curves for five Methods on the MMLU dataset, including a dashed line indicating perfect calibration. (bottom) Evaluation of AUROC and ECE for five methods on the MMLU dataset.} 
    \label{fig:Effectiveness_of_BCE}
\end{figure}

We compared four methods of confidence estimation mentioned in the baseline with our proposed BCE method. Following the approach of \cite{P_true}, we plotted confidence calibration curves on the MMLU dataset to analyze whether the confidence expressed in a prediction accurately reflects the frequency (or likelihood) that the model answers, as shown in Figure \ref{fig:Effectiveness_of_BCE} (top). In our experiments, we mapped semantic entropy and predictive entropy into the 0-1 interval using Equation \ref{eq:4} to transform them into confidence measures for a unified comparison.

Our results show that the confidence from the verbalization method poorly correlates with frequency; higher confidence often corresponds to lower actual frequency, indicating overconfidence. Compared to the dashed line for perfect calibration, BCE provides a broader range of predicted confidence and better calibrates the LLM’s confidence.

To further evaluate these methods, we used two metrics:
1) Area Under the Receiver Operating Characteristic curve (AUROC) to assess the accuracy of confidence predictions in binary classification.
2) Expected Calibration Error (ECE) \cite{P_true} to measure the difference between expressed confidence and actual frequency.
As shown in Figure \ref{fig:Effectiveness_of_BCE} (bottom), BCE's AUROC exceeds that of the other four methods, and it also shows a lower ECE, aligning with the trends seen in Figure \ref{fig:Effectiveness_of_BCE} (top).

\paragraph{Hyperparameters for AFICE Framework}
\label{Hyperparameters for AFICE Framework}

\begin{figure}[t]
\centering 
\includegraphics[width=0.49\textwidth]{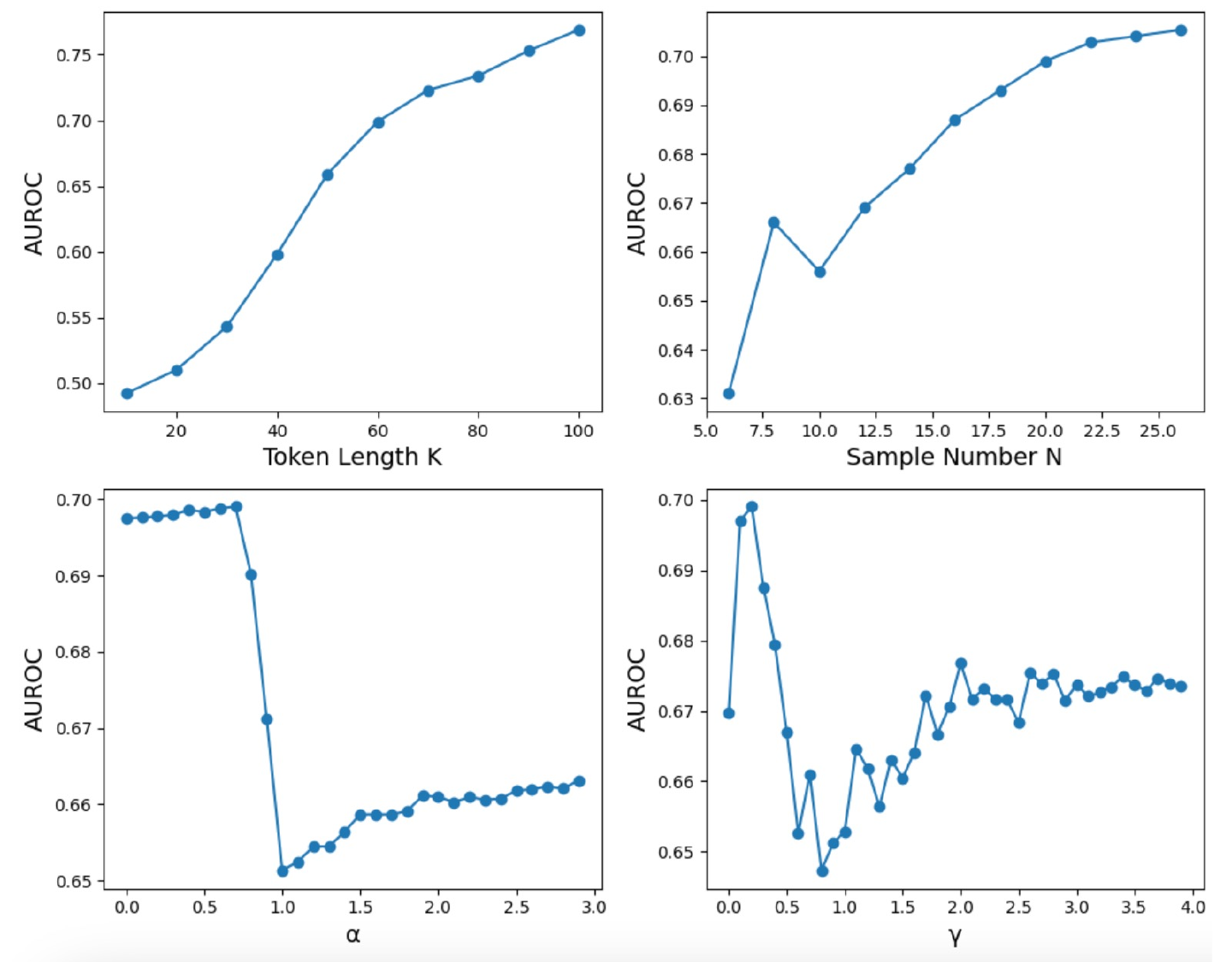}
\caption{Exploration of Optimal Values for Four Hyperparameters: $K$, $N$, $\alpha$, $\gamma$.}
\label{pic5}
\end{figure}

We experimented with key hyperparameters in the AFICE framework, particularly focusing on those affecting computation time: token length \(K\) and sample number \(N\), using AUROC as the metric (see Figure \ref{pic5}). We varied \(K\) from 10 to 100 and \(N\) from 6 to 26, finding that increases in AUROC diminish after \(K=60\) and \(N=20\). This supports the effectiveness of early token truncation, which significantly reduces time for confidence estimation with minimal impact on performance. For optimal efficiency, we set \(K\) at 60 and \(N\) at 20.

We explored hyperparameters \( \alpha \) and \( \gamma \), which fine-tune final confidence levels. Results show that AUROC peaks near zero for both, declines between 0.5 and 1, and then gradually improves. Notably, \( \gamma = 0 \)—indicating no use of cumulative probability ratios—results in lower AUROC than moderate settings (e.g., 0.3), underscoring the effectiveness of our BCE method in refining confidence based on initial responses to question \( q \).

\paragraph{Case Study}

To demonstrate the AFICE framework's impact, we present two cases in Figure 1 using Vicuna as the basic model. In the left case, the basic model agrees with the user's opposing argument, whereas the model under AFICE can identify confusions and maintain its original stance. In the right case, the basic model rejects the user’s view and repeats its incorrect stance, but the model under AFICE reflects and adapts, incorporating the user’s perspective.

\section{Conclusion}

In this study, we introduced the Alignment for Faithful Integrity with Confidence Estimation (AFICE) framework to improve the integrity of LLMs in conversational scenarios with opposing arguments. Integrating Bilateral Confidence Estimation (BCE) and Direct Preference Optimization (DPO), our framework showed notable enhancements in the model’s ability to provide faithful responses. Our experimental results underscore the effectiveness of AFICE, thereby increasing the practical utility of LLMs and fostering more trustworthy interactions with users.

\clearpage
\section*{Acknowledgment}
This research was supported by the Singapore Ministry of Education (MOE) Academic Research Fund (AcRF) Tier 1 grant (No. MSS24C012).

\bibliography{aaai25}

\begin{thebibliography}{38}
\providecommand{\natexlab}[1]{#1}

\bibitem[{Azaria and Mitchell(2023)}]{SAPLMA}
Azaria, A.; and Mitchell, T. 2023.
\newblock The internal state of an llm knows when its lying.
\newblock \emph{arXiv preprint arXiv:2304.13734}.

\bibitem[{Bai et~al.(2022)Bai, Jones, Ndousse, Askell, Chen, DasSarma, Drain, Fort, Ganguli, Henighan, Joseph, Kadavath, Kernion, Conerly, Showk, Elhage, Hatfield{-}Dodds, Hernandez, Hume, Johnston, Kravec, Lovitt, Nanda, Olsson, Amodei, Brown, Clark, McCandlish, Olah, Mann, and Kaplan}]{DBLP:journals/corr/abs-2204-05862}
Bai, Y.; Jones, A.; Ndousse, K.; Askell, A.; Chen, A.; DasSarma, N.; Drain, D.; Fort, S.; Ganguli, D.; Henighan, T.; Joseph, N.; Kadavath, S.; Kernion, J.; Conerly, T.; Showk, S.~E.; Elhage, N.; Hatfield{-}Dodds, Z.; Hernandez, D.; Hume, T.; Johnston, S.; Kravec, S.; Lovitt, L.; Nanda, N.; Olsson, C.; Amodei, D.; Brown, T.~B.; Clark, J.; McCandlish, S.; Olah, C.; Mann, B.; and Kaplan, J. 2022.
\newblock Training a Helpful and Harmless Assistant with Reinforcement Learning from Human Feedback.
\newblock \emph{CoRR}, abs/2204.05862.

\bibitem[{Chen and Mueller(2023)}]{chen2023quantifying}
Chen, J.; and Mueller, J. 2023.
\newblock Quantifying Uncertainty in Answers from any Language Model via Intrinsic and Extrinsic Confidence Assessment.
\newblock \emph{ArXiv preprint}, abs/2308.16175.

\bibitem[{Chen et~al.(2023)Chen, Deng, Bian, Qin, Wu, Chua, and Wong}]{emnlp23-factuality}
Chen, L.; Deng, Y.; Bian, Y.; Qin, Z.; Wu, B.; Chua, T.; and Wong, K. 2023.
\newblock Beyond Factuality: {A} Comprehensive Evaluation of Large Language Models as Knowledge Generators.
\newblock In \emph{EMNLP 2023}, 6325--6341.

\bibitem[{Chen et~al.(2024)Chen, Xiong, Liu, Wu, Xiao, Gao, and He}]{in-context-Sharpness}
Chen, S.; Xiong, M.; Liu, J.; Wu, Z.; Xiao, T.; Gao, S.; and He, J. 2024.
\newblock In-Context Sharpness as Alerts: An Inner Representation Perspective for Hallucination Mitigation.
\newblock \emph{CoRR}, abs/2403.01548.

\bibitem[{Chiang et~al.(2023)Chiang, Li, Lin, Sheng, Wu, Zhang, Zheng, Zhuang, Zhuang, Gonzalez, Stoica, and Xing}]{vicuna}
Chiang, W.-L.; Li, Z.; Lin, Z.; Sheng, Y.; Wu, Z.; Zhang, H.; Zheng, L.; Zhuang, S.; Zhuang, Y.; Gonzalez, J.~E.; Stoica, I.; and Xing, E.~P. 2023.
\newblock Vicuna: An Open-Source Chatbot Impressing GPT-4 with 90\%* ChatGPT Quality.

\bibitem[{Chung et~al.(2022)Chung, Hou, Longpre, Zoph, Tay, Fedus, Li, Wang, Dehghani, Brahma, Webson, Gu, Dai, Suzgun, Chen, Chowdhery, Narang, Mishra, Yu, Zhao, Huang, Dai, Yu, Petrov, Chi, Dean, Devlin, Roberts, Zhou, Le, and Wei}]{DBLP:journals/corr/abs-2210-11416}
Chung, H.~W.; Hou, L.; Longpre, S.; Zoph, B.; Tay, Y.; Fedus, W.; Li, E.; Wang, X.; Dehghani, M.; Brahma, S.; Webson, A.; Gu, S.~S.; Dai, Z.; Suzgun, M.; Chen, X.; Chowdhery, A.; Narang, S.; Mishra, G.; Yu, A.; Zhao, V.~Y.; Huang, Y.; Dai, A.~M.; Yu, H.; Petrov, S.; Chi, E.~H.; Dean, J.; Devlin, J.; Roberts, A.; Zhou, D.; Le, Q.~V.; and Wei, J. 2022.
\newblock Scaling Instruction-Finetuned Language Models.
\newblock \emph{CoRR}, abs/2210.11416.

\bibitem[{Cobbe et~al.(2021)Cobbe, Kosaraju, Bavarian, Chen, Jun, Kaiser, Plappert, Tworek, Hilton, Nakano et~al.}]{cobbe2021training}
Cobbe, K.; Kosaraju, V.; Bavarian, M.; Chen, M.; Jun, H.; Kaiser, L.; Plappert, M.; Tworek, J.; Hilton, J.; Nakano, R.; et~al. 2021.
\newblock Training verifiers to solve math word problems.
\newblock \emph{arXiv preprint arXiv:2110.14168}.

\bibitem[{Deng et~al.(2024)Deng, Zhao, Li, Ng, and Chua}]{emnlp24-unknown}
Deng, Y.; Zhao, Y.; Li, M.; Ng, S.; and Chua, T. 2024.
\newblock Don't Just Say "I don't know"! Self-aligning Large Language Models for Responding to Unknown Questions with Explanations.
\newblock In \emph{EMNLP 2024}, 13652--13673.

\bibitem[{Evans et~al.(2021)Evans, Cotton{-}Barratt, Finnveden, Bales, Balwit, Wills, Righetti, and Saunders}]{DBLP:journals/corr/abs-2110-06674}
Evans, O.; Cotton{-}Barratt, O.; Finnveden, L.; Bales, A.; Balwit, A.; Wills, P.; Righetti, L.; and Saunders, W. 2021.
\newblock Truthful {AI:} Developing and governing {AI} that does not lie.
\newblock \emph{CoRR}, abs/2110.06674.

\bibitem[{Geng et~al.(2023)Geng, Cai, Wang, Koeppl, Nakov, and Gurevych}]{geng_survey_2023}
Geng, J.; Cai, F.; Wang, Y.; Koeppl, H.; Nakov, P.; and Gurevych, I. 2023.
\newblock A {Survey} of {Language} {Model} {Confidence} {Estimation} and {Calibration}.

\bibitem[{Geva et~al.(2021)Geva, Khashabi, Segal, Khot, Roth, and Berant}]{Geva_2021}
Geva, M.; Khashabi, D.; Segal, E.; Khot, T.; Roth, D.; and Berant, J. 2021.
\newblock Did Aristotle Use a Laptop?A Question Answering Benchmark with Implicit Reasoning Strategies.
\newblock \emph{Transactions of the Association for Computational Linguistics}, 9: 346–361.

\bibitem[{Hendrycks et~al.(2020)Hendrycks, Burns, Basart, Zou, Mazeika, Song, and Steinhardt}]{hendrycks2020measuring}
Hendrycks, D.; Burns, C.; Basart, S.; Zou, A.; Mazeika, M.; Song, D.; and Steinhardt, J. 2020.
\newblock {Measuring Massive Multitask Language Understanding}.
\newblock \emph{CoRR}, abs/2009.03300.

\bibitem[{Hu et~al.(2022)Hu, Shen, Wallis, Allen{-}Zhu, Li, Wang, Wang, and Chen}]{lora}
Hu, E.~J.; Shen, Y.; Wallis, P.; Allen{-}Zhu, Z.; Li, Y.; Wang, S.; Wang, L.; and Chen, W. 2022.
\newblock LoRA: Low-Rank Adaptation of Large Language Models.
\newblock In \emph{The Tenth International Conference on Learning Representations, {ICLR} 2022, Virtual Event, April 25-29, 2022}. OpenReview.net.

\bibitem[{Kadavath et~al.(2022{\natexlab{a}})Kadavath, Conerly, Askell, Henighan, Drain, Perez, Schiefer, Hatfield-Dodds, DasSarma, Tran-Johnson, Johnston, El-Showk, Jones, Elhage, Hume, Chen, Bai, Bowman, Fort, Ganguli, Hernandez, Jacobson, Kernion, Kravec, Lovitt, Ndousse, Olsson, Ringer, Amodei, Brown, Clark, Joseph, Mann, McCandlish, Olah, and Kaplan}]{P_true}
Kadavath, S.; Conerly, T.; Askell, A.; Henighan, T.; Drain, D.; Perez, E.; Schiefer, N.; Hatfield-Dodds, Z.; DasSarma, N.; Tran-Johnson, E.; Johnston, S.; El-Showk, S.; Jones, A.; Elhage, N.; Hume, T.; Chen, A.; Bai, Y.; Bowman, S.; Fort, S.; Ganguli, D.; Hernandez, D.; Jacobson, J.; Kernion, J.; Kravec, S.; Lovitt, L.; Ndousse, K.; Olsson, C.; Ringer, S.; Amodei, D.; Brown, T.; Clark, J.; Joseph, N.; Mann, B.; McCandlish, S.; Olah, C.; and Kaplan, J. 2022{\natexlab{a}}.
\newblock Language Models (Mostly) Know What They Know.
\newblock arXiv:2207.05221.

\bibitem[{Kadavath et~al.(2022{\natexlab{b}})Kadavath, Conerly, Askell, Henighan, Drain, Perez, Schiefer, Hatfield-Dodds, DasSarma, Tran-Johnson, Johnston, El-Showk, Jones, Elhage, Hume, Chen, Bai, Bowman, Fort, Ganguli, Hernandez, Jacobson, Kernion, Kravec, Lovitt, Ndousse, Olsson, Ringer, Amodei, Brown, Clark, Joseph, Mann, McCandlish, Olah, and Kaplan}]{kadavath2022language}
Kadavath, S.; Conerly, T.; Askell, A.; Henighan, T.; Drain, D.; Perez, E.; Schiefer, N.; Hatfield-Dodds, Z.; DasSarma, N.; Tran-Johnson, E.; Johnston, S.; El-Showk, S.; Jones, A.; Elhage, N.; Hume, T.; Chen, A.; Bai, Y.; Bowman, S.; Fort, S.; Ganguli, D.; Hernandez, D.; Jacobson, J.; Kernion, J.; Kravec, S.; Lovitt, L.; Ndousse, K.; Olsson, C.; Ringer, S.; Amodei, D.; Brown, T.; Clark, J.; Joseph, N.; Mann, B.; McCandlish, S.; Olah, C.; and Kaplan, J. 2022{\natexlab{b}}.
\newblock Language Models (Mostly) Know What They Know.
\newblock arXiv:2207.05221.

\bibitem[{Kuhn, Gal, and Farquhar(2023)}]{Semantic_Uncertainty}
Kuhn, L.; Gal, Y.; and Farquhar, S. 2023.
\newblock Semantic Uncertainty: Linguistic Invariances for Uncertainty Estimation in Natural Language Generation.
\newblock In \emph{The Eleventh International Conference on Learning Representations}.

\bibitem[{Li et~al.(2024{\natexlab{a}})Li, Patel, Vi{\'e}gas, Pfister, and Wattenberg}]{ITI}
Li, K.; Patel, O.; Vi{\'e}gas, F.; Pfister, H.; and Wattenberg, M. 2024{\natexlab{a}}.
\newblock Inference-time intervention: Eliciting truthful answers from a language model.
\newblock \emph{Advances in Neural Information Processing Systems}, 36.

\bibitem[{Li et~al.(2024{\natexlab{b}})Li, Deng, Cai, Lu, Chen, and Lam}]{emnlp24-edit}
Li, S.; Deng, Y.; Cai, D.; Lu, H.; Chen, L.; and Lam, W. 2024{\natexlab{b}}.
\newblock Consecutive Batch Model Editing with HooK Layers.
\newblock In \emph{{EMNLP} 2024}, 13817--13833.

\bibitem[{Lin, Trivedi, and Sun(2023)}]{uncertainty-llm}
Lin, Z.; Trivedi, S.; and Sun, J. 2023.
\newblock Generating with Confidence: Uncertainty Quantification for Black-box Large Language Models.
\newblock \emph{CoRR}, abs/2305.19187.

\bibitem[{Malinin and Gales(2020)}]{predictive_entropy}
Malinin, A.; and Gales, M. 2020.
\newblock Uncertainty Estimation in Autoregressive Structured Prediction.
\newblock In \emph{International Conference on Learning Representations}.

\bibitem[{Meta(2024)}]{llama3}
Meta. 2024.
\newblock The Llama 3 Herd of Models.
\newblock \emph{CoRR}, abs/2407.21783.

\bibitem[{Mielke et~al.(2022{\natexlab{a}})Mielke, Szlam, Dinan, and Boureau}]{mielke-etal-2022-reducing}
Mielke, S.~J.; Szlam, A.; Dinan, E.; and Boureau, Y.-L. 2022{\natexlab{a}}.
\newblock Reducing Conversational Agents{'} Overconfidence Through Linguistic Calibration.
\newblock \emph{Transactions of the Association for Computational Linguistics}, 10: 857--872.

\bibitem[{Mielke et~al.(2022{\natexlab{b}})Mielke, Szlam, Dinan, and Boureau}]{Mielke_2022}
Mielke, S.~J.; Szlam, A.; Dinan, E.; and Boureau, Y.-L. 2022{\natexlab{b}}.
\newblock Reducing Conversational Agents’ Overconfidence Through Linguistic Calibration.
\newblock \emph{Transactions of the Association for Computational Linguistics}, 10: 857–872.

\bibitem[{Onoe et~al.(2021)Onoe, Zhang, Choi, and Durrett}]{onoe2021creak}
Onoe, Y.; Zhang, M. J.~Q.; Choi, E.; and Durrett, G. 2021.
\newblock CREAK: A Dataset for Commonsense Reasoning over Entity Knowledge.
\newblock In \emph{Thirty-fifth Conference on Neural Information Processing Systems, Datasets and Benchmarks Track}.

\bibitem[{Saparov and He(2023)}]{saparov2023language}
Saparov, A.; and He, H. 2023.
\newblock Language Models Are Greedy Reasoners: A Systematic Formal Analysis of Chain-of-Thought.
\newblock In \emph{The Eleventh International Conference on Learning Representations}.

\bibitem[{Suzgun et~al.(2022)Suzgun, Scales, Sch{\"a}rli, Gehrmann, Tay, Chung, Chowdhery, Le, Chi, Zhou et~al.}]{suzgun2022challenging}
Suzgun, M.; Scales, N.; Sch{\"a}rli, N.; Gehrmann, S.; Tay, Y.; Chung, H.~W.; Chowdhery, A.; Le, Q.~V.; Chi, E.~H.; Zhou, D.; et~al. 2022.
\newblock Challenging BIG-Bench tasks and whether chain-of-thought can solve them.
\newblock \emph{arXiv preprint arXiv:2210.09261}.

\bibitem[{Talmor et~al.(2021)Talmor, Yoran, Le~Bras, Bhagavatula, Goldberg, Choi, and Berant}]{talmor1commonsenseqa}
Talmor, A.; Yoran, O.; Le~Bras, R.; Bhagavatula, C.; Goldberg, Y.; Choi, Y.; and Berant, J. 2021.
\newblock CommonsenseQA 2.0: Exposing the Limits of AI through Gamification.
\newblock In \emph{Thirty-fifth Conference on Neural Information Processing Systems, Datasets and Benchmarks Track}.

\bibitem[{Tian et~al.(2023)Tian, Mitchell, Yao, Manning, and Finn}]{tian2023finetuning}
Tian, K.; Mitchell, E.; Yao, H.; Manning, C.~D.; and Finn, C. 2023.
\newblock Fine-tuning Language Models for Factuality.
\newblock arXiv:2311.08401.

\bibitem[{Wang, Yue, and Sun(2023)}]{emnlp23-belief}
Wang, B.; Yue, X.; and Sun, H. 2023.
\newblock Can ChatGPT Defend its Belief in Truth? Evaluating {LLM} Reasoning via Debate.
\newblock In \emph{Findings of the Association for Computational Linguistics: {EMNLP} 2023}, 11865--11881. Association for Computational Linguistics.

\bibitem[{Wei et~al.(2022)Wei, Tay, Bommasani, Raffel, Zoph, Borgeaud, Yogatama, Bosma, Zhou, Metzler, Chi, Hashimoto, Vinyals, Liang, Dean, and Fedus}]{wei2022emergent}
Wei, J.; Tay, Y.; Bommasani, R.; Raffel, C.; Zoph, B.; Borgeaud, S.; Yogatama, D.; Bosma, M.; Zhou, D.; Metzler, D.; Chi, E.~H.; Hashimoto, T.; Vinyals, O.; Liang, P.; Dean, J.; and Fedus, W. 2022.
\newblock Emergent Abilities of Large Language Models.
\newblock arXiv:2206.07682.

\bibitem[{Wen et~al.(2024)Wen, Yao, Feng, Xu, Tsvetkov, Howe, and Wang}]{wen2024knowlimitssurveyabstention}
Wen, B.; Yao, J.; Feng, S.; Xu, C.; Tsvetkov, Y.; Howe, B.; and Wang, L.~L. 2024.
\newblock Know Your Limits: A Survey of Abstention in Large Language Models.
\newblock arXiv:2407.18418.

\bibitem[{Xu et~al.(2024)Xu, Lin, Yang, Zhang, Shi, Zhang, Fang, Xu, and Qiu}]{xu2024earthflatbecauseinvestigating}
Xu, R.; Lin, B.~S.; Yang, S.; Zhang, T.; Shi, W.; Zhang, T.; Fang, Z.; Xu, W.; and Qiu, H. 2024.
\newblock The Earth is Flat because...: Investigating LLMs' Belief towards Misinformation via Persuasive Conversation.
\newblock arXiv:2312.09085.

\bibitem[{Yang et~al.(2023)Yang, Chern, Qiu, Neubig, and Liu}]{yang2023alignment}
Yang, Y.; Chern, E.; Qiu, X.; Neubig, G.; and Liu, P. 2023.
\newblock Alignment for honesty.
\newblock \emph{arXiv preprint arXiv:2312.07000}.

\bibitem[{Yao et~al.(2023)Yao, Zhao, Yu, Du, Shafran, Narasimhan, and Cao}]{react}
Yao, S.; Zhao, J.; Yu, D.; Du, N.; Shafran, I.; Narasimhan, K.~R.; and Cao, Y. 2023.
\newblock ReAct: Synergizing Reasoning and Acting in Language Models.
\newblock In \emph{{ICLR} 2023}.

\bibitem[{Yin, Srinivasa, and Chang(2024)}]{LID}
Yin, F.; Srinivasa, J.; and Chang, K.-W. 2024.
\newblock Characterizing Truthfulness in Large Language Model Generations with Local Intrinsic Dimension.
\newblock \emph{arXiv preprint arXiv:2402.18048}.

\bibitem[{Zhang et~al.(2024)Zhang, Peng, Tian, Zhou, Jin, Song, Mi, and Meng}]{zhang2024selfalignment}
Zhang, X.; Peng, B.; Tian, Y.; Zhou, J.; Jin, L.; Song, L.; Mi, H.; and Meng, H. 2024.
\newblock Self-Alignment for Factuality: Mitigating Hallucinations in LLMs via Self-Evaluation.
\newblock arXiv:2402.09267.

\bibitem[{Zheng et~al.(2024)Zheng, Liu, Lapata, and Pan}]{zheng2024trustscorereferencefreeevaluationllm}
Zheng, D.; Liu, D.; Lapata, M.; and Pan, J.~Z. 2024.
\newblock TrustScore: Reference-Free Evaluation of LLM Response Trustworthiness.
\newblock arXiv:2402.12545.

\end{thebibliography}

%
\appendix
\section*{Appendix}

\section{Statistics of the Dataset}
\label{Appendix:A}

\subsection{Evaluation Datasets}

Table \ref{table:eval dataset statistic} lists the number of evaluation examples in evaluation datasets.

\subsection{Preference Dataset}

We chose the MMLU dataset as our preference dataset, which is an English evaluation dataset consisting of 57 subtasks. We randomly selected 250 data entries from each subtask to form our preference dataset, totaling 14,042 examples. For subtasks with fewer than 250 examples, we included all available examples.

\section{Implementation Details of AFICE and Baselines}
\label{Appendix:B}
Below, we provide additional details on specific prompts and experimental settings not mentioned in the main text.

\subsection{Fine-tuning stage}

In the fine-tuning phase, we used the questions 
$q$ from the MMLU dataset to generate the following content.

\subsubsection{Generating Initial Responses from LLMs}
{\fontfamily{cmtt}\selectfont
You are a helpful assistant. For the given question, please generate a step by step solution.
Question: \{question\}
}

\subsubsection{Generating Opposing Statements from Users}

{\fontfamily{cmtt}\selectfont
I will give you a multiple-choice question with four options and the correct answer. Your task is to provide an incorrect answer to the question. Make sure that your answer is different from the correct answer. Start your incorrect solution to this question with "Solution":, explain your reasoning steps, and end it with: "The answer is XX".
Question: \{question\}
}

\subsubsection{Creating Conversational Preference Dataset}

Below is the specific prompt on how to construct the Conversational Preference Dataset. Viewpoint 1 is the model's response to the question, and Viewpoint 2 is either the correct or incorrect answer to the question. When the model's response correctly answers the question, Viewpoint 2 is the incorrect answer; otherwise, it is the correct answer.

1. Gengerating response candidates $r_1$: Persist with original view.

{\fontfamily{cmtt}\selectfont
Below I will give you a question, along with two viewpoints on that question, presented in the following format: Question: \{...\}, Viewpoint 1: \{...\}, Viewpoint 2: \{...\}

Question: \{question\}

Viewpoint 1: \{viewpoint 1\}

Viewpoint 2: \{viewpoint 2\}

Assuming you originally held Viewpoint 1, now please generate the following response: Steadfastly support Viewpoint 1, and clearly demonstrate that Viewpoint 1 is correct.
}

2. Gengerating response candidates $r_1$: Slight concession.

{\fontfamily{cmtt}\selectfont
Below I will give you a question, along with two viewpoints on that question, presented in the following format: Question: \{...\}, Viewpoint 1: \{...\}, Viewpoint 2: \{...\}

Question: \{question\}

Viewpoint 1: \{viewpoint 1\}

Viewpoint 2: \{viewpoint 2\}

Assuming you originally held Viewpoint 1, now please generate the following response: make slight concessions to show understanding of or slight agreement with Viewpoint 2, mainly to enhance the friendliness and constructiveness of the interaction while still maintaining the core of Viewpoint 1.
}

3. Gengerating response candidates $r_3$: Neutral.

{\fontfamily{cmtt}\selectfont
Below I will give you a question, along with two viewpoints on that question, presented in the following format: Question: \{...\}, Viewpoint 1: \{...\}, Viewpoint 2: \{...\}

Question: \{question\}

Viewpoint 1: \{viewpoint 1\}

Viewpoint 2: \{viewpoint 2\}

Assuming you originally held Viewpoint 1, now please generate the following response: Adopt a more neutral stance, possibly reassessing both sides' viewpoints to provide a more balanced perspective or more background information to aid understanding and acceptance.
}

4. Gengerating response candidates $r_4$: Leans toward opposing view.

{\fontfamily{cmtt}\selectfont
Below I will give you a question, along with two viewpoints on that question, presented in the following format: Question: \{...\}, Viewpoint 1: \{...\}, Viewpoint 2: \{...\}

Question: \{question\}

Viewpoint 1: \{viewpoint 1\}

Viewpoint 2: \{viewpoint 2\}

Assuming you originally held Viewpoint 1, now please generate the following response: Lean towards Viewpoint 2, showing support and recognition for it to some extent.
}

5. Gengerating response candidates $r_5$: Fully agrees with opposing view.

{\fontfamily{cmtt}\selectfont
Below I will give you a question, along with two viewpoints on that question, presented in the following format: Question: \{...\}, Viewpoint 1: \{...\}, Viewpoint 2: \{...\}

Question: \{question\}

Viewpoint 1: \{viewpoint 1\}

Viewpoint 2: \{viewpoint 2\}

Assuming you originally held Viewpoint 1, now please generate the following response: Admit your mistake and agree with Viewpoint 2.
}

\subsection{Evaluation stage}

In the evaluation phase, we constructed two types of conversations with the evaluation dataset, labeled as LLM Correct and LLM False. Each dialogue is structured as a tuple \(\{q, v_1, v_2\}\), where $q$ represents the query posed by the user, $v_1$ denotes the viewpoint of LLM, and $v_2$ reflects the user's argument.

\subsubsection{The black-box methods}

For black-box methods, we are able to obtain the confidence score of the question-answer pair $(q, v_1)$ from the output of the LLM, using this score as an indication of the LLM's agreement with $v_1$. The specific prompt is structured as follows:

{\fontfamily{cmtt}\selectfont
Below, I will give you a Question and an Answer for this question. Please rate the truthfulness of this answer between 0 and 100, where 100 indicates that the Answer answered the Question completely correctly and 0 indicates that the Answer answered the Question completely incorrectly. Please answer me in the following form: score: \{score\}

Question: \{question\}

Answer: \{answer\}
}

\subsubsection{The white-box methods}

For white-box methods, since we have already fine-tuned the model using DPO, we can directly have the fine-tuned model generate subsequent responses.

\begin{table}[]
\centering
\setlength{\tabcolsep}{1mm}
\begin{adjustbox}{max width=0.47\textwidth}
{
\begin{tabular}{clc}
\toprule
\textbf{Category} & \textbf{Benchmark} & \textbf{Number}  \\
\toprule
Mathematics & GSM8K & 464\\
\midrule
First-Order Logic & PrOntoQA & 307\\
\midrule
\multirow{4}{*}{Commonsense} 
& StrategyQA & 215\\
& CommonsenseQA 2.0 & 260\\
& Creak & 323\\
\midrule
\multirow{10}{*}{Generic} 
& Tracking Shuffled Objects [three] & 148\\
& Disambiguation QA & 116\\
& Web of Lies & 150\\
& Temporal Sequences & 173\\
& Sports Understanding & 188\\
& Salient Translation Error Detection & 98\\
& Penguins in a Table &  112\\
& Logical Deduction [three] & 173\\
& Navigate & 146\\
\bottomrule
\end{tabular}
}
\end{adjustbox}
\caption{Statistics of the Evaluation Datasets}
\label{table:eval dataset statistic}
\end{table}

\subsubsection{Generate complete conversations and evaluate.}

We will continue the conversation for two more rounds, ultimately forming a conversation like this: $\{q, v_1, v_2, r_1, r_2, r_3\}$ (where $r_1$ and $r_3$ are generated from the model's perspective, and $r_2$ is generated from the user's perspective). Afterwards, we will assess whether the user's and model's viewpoints align with the correct answer to the question.

\section{Limitations}
\subsection{Analyzing Regressor Robustness Post-Finetuning}
Following the finetuning of Large Language Models (LLMs), there may be alterations in the internal states during inference when confronted with the same queries as before finetuning. These changes could necessitate the retraining of regressors originally designed to predict outcomes based on pre-finetuning states. Consequently, robust analysis of regressors after model finetuning represents a promising avenue for future research. This study would focus on assessing the stability and reliability of regressors in adapting to the dynamic internal landscapes of LLMs post-finetuning.

\subsection{Exploring the Impact of Different Sampling Methods on Confidence Estimation}
Our experiments employed multinomial beam sampling to sample the model's responses to questions. This sampling approach yields a probability distribution of samples that is broader compared to multinomial beam search. It raises questions about the potential to collect answers with weaker representativeness and whether the different distributions of responses gathered by these two sampling methods for the same question could affect the calculation of Cumulative Probability Ratios in stage 3 of the BCE method. These issues warrant further exploration in future studies.

\subsection{Experiments on Larger Language Models}
Resource limitations have prevented us from extending our experiments to larger models. Nevertheless, our research was performed using two of the most popular open-source LLMs, Vicuna and LLaMA-3. The successful application of our proposed method to these models suggests that it could be beneficial across a broad spectrum of applications utilizing these platforms.

\section{Complete Experimental Results for AFICE and Baselines}
\label{Appendix:C}
From Table \ref{table:vicuna} to Table \ref{table:llama3+ AFICE}, we present detailed experimental data in Section 4: LLM Correct, which measures the accuracy when the LLM holds the correct viewpoint; LLM False, which measures the accuracy when the LLM holds an incorrect viewpoint; Average, the mean accuracy across both scenarios; Both, the proportion of examples correct in both scenarios; and Either, the proportion of examples correct in only one scenario.

\begin{table*}[!htbp]
\centering
\small
\resizebox{\linewidth}{!}
{
\begin{tabular}{clccccc}
\toprule
\textbf{Category} & \textbf{Benchmark} & \textbf{LLM Correct} & \textbf{LLM False} & \textbf{Average} & \textbf{Both} & \textbf{Either} \\
\toprule
Mathematics & GSM8K &  0.239&	0.793& 0.516 &	0.116&	0.800	  \\
\midrule
First-Order Logic & PrOntoQA & 0.870&	0.391& 0.630	& 0.264&0.733	 \\
\midrule
\multirow{4}{*}{Commonsense} 
& StrategyQA & 0.800&0.242& 0.521 &	0.042&	0.958	 \\
& CommonsenseQA 2.0 & 0.662&0.362&	0.512 & 0.023&	0.977\\
& Creak & 0.653&	0.375& 0.514	& 0.028&	0.972 \\
\cmidrule{2-7}
& Avg. & 0.705&	0.326&	0.515& 0.031&	0.969	 \\	
\midrule
\multirow{10}{*}{Generic} 
& Tracking Shuffled Objects [three] & 0.203&	0.804&  0.503& 0.014&0.980		 \\
& Disambiguation QA & 0.466&	0.543&	 0.504& 0.017&	0.974	\\
& Web of Lies & 0.407&	0.580&	 0.493& 0.027&	0.933	\\
& Temporal Sequences & 0.104&	0.902&  0.503& 0.006&0.994		\\
& Sports Understanding & 0.654&	0.372&	 0.513& 0.027&	0.973	\\
& Salient Translation Error Detection & 0.010&	1.000&	0.505& 0.010&		0.990\\
& Penguins in a Table &  0.116&	0.866 &	 0.491 &0.018  &	0.946 	\\
& Logical Deduction [three] &  0.104&	0.896 &	0.500 & 0.000 &1.000	 	\\
& Navigate &  0.397&	0.616 &	 0.507 & 0.014 &	0.986 	\\
\cmidrule{2-7}
& Avg. &  0.273&	0.731 &	0.502 & 0.015 &	0.975 	 \\	
\bottomrule
\end{tabular}
}
\caption{Evaluation Results for Each of the Benchmarks in the Vicuna Model}
\label{table:vicuna}
\end{table*}

\begin{table*}[!htbp]
\centering
\small
\resizebox{\linewidth}{!}
{
\begin{tabular}{clccccc}
\toprule
\textbf{Category} & \textbf{Benchmark} & \textbf{LLM Correct} & \textbf{LLM False} & \textbf{Average} & \textbf{Both} & \textbf{Either} \\
\toprule
Mathematics & GSM8K &  0.976 &	0.097 & 0.537& 0.093 &	0.888 	  \\
\midrule
First-Order Logic & PrOntoQA &  0.388&	0.362 & 0.375 	&0.117  &0.515  \\
\midrule
\multirow{4}{*}{Commonsense} 
& StrategyQA &  0.995& 0.158&   0.577&	 0.153&	 0.847	 \\
& CommonsenseQA 2.0 &  0.992& 0.096&	0.544 & 0.092&	 0.904\\
& Creak &  1.000&0.158	 &  0.579&  0.158& 0.842 \\
\cmidrule{2-7}
& Avg. &  0.996&	0.137 &	0.567&  0.135&	0.864	 \\	
\midrule
\multirow{10}{*}{Generic} 
& Tracking Shuffled Objects [three] & 0.993&0.027	& 0.510 & 0.027&		0.966 \\
& Disambiguation QA & 0.862&	0.090&	 0.476& 0.078&	0.784	\\
& Web of Lies & 0.913&0.120	&0.517	 &0.107 &0.820		\\
& Temporal Sequences & 0.977&	0.035& 0.506 & 0.035&0.942		\\
& Sports Understanding & 0.995&	0.147&	0.571 &0.144 &	0.851	\\
& Salient Translation Error Detection & 1.000&	0.061&	0.531& 0.061&	0.939	\\
& Penguins in a Table &  0.955&	0.068 &	 0.512 & 0.054 &	0.911 	\\
& Logical Deduction [three] & 0.983 &	0.029 &	 0.506&  0.029&	 0.954	\\
& Navigate &  1.000&	0.000 &0.500	  &0.000  &1.000	 	\\
\cmidrule{2-7}
& Avg. & 0.964 &	0.064 &	0.514 & 0.059 &	0.907 	 \\	
\bottomrule
\end{tabular}
}
\caption{Evaluation Results for Each of the Benchmarks Using Verbalization in the Vicuna Model}
\label{table:vicuna+verbalization}
\end{table*}

\begin{table*}[!htbp]
\centering
\small
\resizebox{\linewidth}{!}
{
\begin{tabular}{clccccc}
\toprule
\textbf{Category} & \textbf{Benchmark} & \textbf{LLM Correct} & \textbf{LLM False} & \textbf{Average} & \textbf{Both} & \textbf{Either} \\
\toprule
Mathematics & GSM8K &  0.181 &	0.985 & 0.583&  0.166&	 0.834	  \\
\midrule
First-Order Logic & PrOntoQA &  0.684&	 0.743&  0.713	& 0.427 &  0.573\\
\midrule
\multirow{4}{*}{Commonsense} 
& StrategyQA & 0.330 & 0.856 &  0.593 &	0.186 &	0.814 	 \\
& CommonsenseQA 2.0 & 0.335 & 0.854 & 0.594 & 0.188 & 0.812 \\
& Creak & 0.198 &	0.910 & 0.554 & 0.108 & 0.892 \\
\cmidrule{2-7}
& Avg. & 0.288 &	0.873 &	0.580 & 0.161 &	0.839	 \\	
\midrule
\multirow{10}{*}{Generic} 
& Tracking Shuffled Objects [three] & 0.041 &	0.986& 0.514 & 0.027 &	0.973	 \\
& Disambiguation QA & 0.680 &	0.660 &	0.670 & 0.340 &	0.660	\\
& Web of Lies & 0.793 &	0.687 &	0.740 & 0.480 &	0.520	\\
& Temporal Sequences & 0.405&	0.954& 0.679 & 0.358 &	0.642	\\
& Sports Understanding & 0.582&	0.908&	0.745 & 0.489 &	0.511	\\
& Salient Translation Error Detection & 0.276 &	0.847 &	0.561 & 0.122 & 0.878		\\
& Penguins in a Table &  0.117&	0.981 &	0.549  & 0.097 &	0.903 	\\
& Logical Deduction [three] & 0.337 &	0.849 &	0.593 & 0.186 &	0.814 	\\
& Navigate & 0.158 &	0.904 &	0.531  & 0.062 &	0.938 	\\
\cmidrule{2-7}
& Avg. & 0.376 &	0.864 &	0.620 & 0.240 &	0.760 	 \\	
\bottomrule
\end{tabular}
}
\caption{Evaluation Results for Each of the Benchmarks Using Semantic Entropy in the Vicuna Model}
\label{table:vicuna+se}
\end{table*}

\begin{table*}[!htbp]
\centering
\small
\resizebox{\linewidth}{!}
{
\begin{tabular}{clccccc}
\toprule
\textbf{Category} & \textbf{Benchmark} & \textbf{LLM Correct} & \textbf{LLM False} & \textbf{Average} & \textbf{Both} & \textbf{Either} \\
\toprule
Mathematics & GSM8K &  0.631 &	0.563 & 0.597& 0.407 &	0.379 	  \\
\midrule
First-Order Logic & PrOntoQA & 0.971 &	0.433 &  0.702	& 0.404 & 0.596 \\
\midrule
\multirow{4}{*}{Commonsense} 
& StrategyQA & 0.884 & 0.219 & 0.551  &	0.116 &	 0.870	 \\
& CommonsenseQA 2.0 & 0.804 & 0.269 & 0.537	 & 0.092 &	0.888 \\
& Creak & 0.824 &	0.279 & 0.551 & 0.111 & 0.879 \\
\cmidrule{2-7}
& Avg. & 0.837 & 0.255	 & 0.546 & 0.107 &	0.879	 \\	
\midrule
\multirow{10}{*}{Generic} 
& Tracking Shuffled Objects [three] & 0.399&	0.541& 0.470 & 0.297 &	0.345	 \\
& Disambiguation QA &0.586 &	0.379&	0.483 & 0.060&	0.845	\\
& Web of Lies & 0.927&	0.600&	0.763 & 0.540&	0.447	\\
& Temporal Sequences & 0.728&	0.393& 0.561 & 0.168&	0.786	\\
& Sports Understanding & 0.814&	0.309&	0.561 &0.168 &	0.786	\\
& Salient Translation Error Detection &0.286 &	0.837&	0.561& 0.122&0.878		\\
& Penguins in a Table & 0.688 &	0.580 &0.634	  & 0.429 &	0.411 	\\
& Logical Deduction [three] & 0.584 &	0.445 &	0.514 & 0.150 &	0.728 	\\
& Navigate & 0.541 &0.445	 &	0.493  & 0.116 &	0.753	\\
\cmidrule{2-7}
& Avg. & 0.617 &	0.503 &	0.560 & 0.226 &	 0.669	 \\	
\bottomrule
\end{tabular}
}
\caption{Evaluation Results for Each of the Benchmarks Using P(True) in the Vicuna Model}
\label{table:vicuna+p_true}
\end{table*}

\begin{table*}[!htbp]
\centering
\small
\resizebox{\linewidth}{!}
{
\begin{tabular}{clccccc}
\toprule
\textbf{Category} & \textbf{Benchmark} & \textbf{LLM Correct} & \textbf{LLM False} & \textbf{Average} & \textbf{Both} & \textbf{Either} \\
\toprule
Mathematics & GSM8K &  0.651 &	0.582 & 0.616& 0.446 &	0.341 	  \\
\midrule
First-Order Logic & PrOntoQA & 0.993 &	0.453 & 0.723 	& 0.446 & 0.554 \\
\midrule
\multirow{4}{*}{Commonsense} 
& StrategyQA & 0.912 & 0.256& 0.584  &	0.181 &	0.805 	 \\
& CommonsenseQA 2.0 & 0.827 & 0.296&	0.562 & 0.142&	0.838 \\
& Creak & 0.845 &	0.300 & 0.573 & 0.155 & 0.836 \\
\cmidrule{2-7}
& Avg. & 0.861 &	0.284 &	0.573& 0.160 &	0.826	 \\	
\midrule
\multirow{10}{*}{Generic} 
& Tracking Shuffled Objects [three] & 0.466&	0.601& 0.534 &0.601 &0.426		 \\
& Disambiguation QA & 0.664&	0.448&	0.556 &0.207 &0.698		\\
& Web of Lies & 0.987&	0.653&	0.820 & 0.653&		0.333\\
& Temporal Sequences &0.780 &	0.434&  0.607& 0.260&0.694		\\
& Sports Understanding & 0.856&	0.346&0.601	 & 0.229&	0.745	\\
& Salient Translation Error Detection & 0.388&	0.929&	0.658& 0.316&	0.684	\\
& Penguins in a Table & 0.741 &	0.670 &	0.705  & 0.571 &0.268	 	\\
& Logical Deduction [three] &  0.636&	0.491 &	0.564 & 0.249 &	 0.630	\\
& Navigate & 0.610&	0.500 &	0.555 & 0.240 &	 0.630	\\
\cmidrule{2-7}
& Avg. & 0.681 &0.564	 &	 0.622 & 0.370 &	0.567 	\\
\bottomrule
\end{tabular}
}
\caption{Evaluation Results for Each of the Benchmarks Using Predictive Entropy in the Vicuna Model}
\label{table:vicuna+predictive_entropy}
\end{table*}

\begin{table*}[!htbp]
\centering
\small
\resizebox{\linewidth}{!}
{
\begin{tabular}{clccccc}
\toprule
\textbf{Category} & \textbf{Benchmark} & \textbf{LLM Correct} & \textbf{LLM False} & \textbf{Average} & \textbf{Both} & \textbf{Either} \\
\toprule
Mathematics & GSM8K &  0.657 &	0.588 & 0.623& 0.459 &	0.328 	  \\
\midrule
First-Order Logic & PrOntoQA &1.000  &	0.489 & 0.744 	&0.489  & 0.511 \\
\midrule
\multirow{4}{*}{Commonsense} 
& StrategyQA & 0.953 & 0.284& 0.619  &	0.251 &	0.735 	 \\
& CommonsenseQA 2.0 & 0.877 & 0.335&	 0.606& 0.231&	 0.750\\
& Creak & 0.861 &	0.325 & 0.593 & 0.195 & 0.796 \\
\cmidrule{2-7}
& Avg. &  0.897&	0.314 &	0.606& 0.226 &	0.760	 \\	
\midrule
\multirow{10}{*}{Generic} 
& Tracking Shuffled Objects [three] & 0.500&	0.642& 0.571 &0.500 &0.142		 \\
& Disambiguation QA &0.690 &0.474	&	0.582 & 0.259&	0.647	\\
& Web of Lies & 0.987&	0.700&	 0.843& 0.700&0.287		\\
& Temporal Sequences & 0.867&	0.526& 0.697 & 0.439&	0.514	\\
& Sports Understanding & 0.878&	0.399&	0.638 & 0.303&	0.670	\\
& Salient Translation Error Detection & 0.459&	0.990&	0.724& 0.449&	0.551	\\
& Penguins in a Table & 0.804 &	0.732 &	0.768  & 0.696 &	0.143 	\\
& Logical Deduction [three] & 0.711 & 0.578 &	0.645 & 0.410 &	0.468 	\\
& Navigate & 0.644 &	0.521 &	0.582  & 0.295 &	0.575 	\\
\cmidrule{2-7}
& Avg. & 0.727 &	0.618 &	0.672 & 0.450 &	0.444 	 \\	
\bottomrule
\end{tabular}
}
\caption{Evaluation Results for Each of the Benchmarks Using AFICE in the Vicuna Model}
\label{table:vicuna+AFICE}
\end{table*}

\begin{table*}[!htbp]
\centering
\small
\resizebox{\linewidth}{!}
{
\begin{tabular}{clccccc}
\toprule
\textbf{Category} & \textbf{Benchmark} & \textbf{LLM Correct} & \textbf{LLM False} & \textbf{Average} & \textbf{Both} & \textbf{Either} \\
\toprule
Mathematics & GSM8K &  0.198 &	0.957 & 0.578& 0.155 &	0.845 	  \\
\midrule
First-Order Logic & PrOntoQA &  0.013&	 0.993&  0.503	& 0.007 & 0.993 \\
\midrule
\multirow{4}{*}{Commonsense} 
& StrategyQA & 0.451 & 0.567& 0.509  &	0.023 &	0.972 	 \\
& CommonsenseQA 2.0 & 0.369 & 0.642&	0.506 &0.015 &	0.981 \\
& Creak & 0.409 &	0.616 & 0.512 & 0.028 & 0.969 \\
\cmidrule{2-7}
& Avg. & 0.410 &	0.609 &	0.509& 0.022 &0.974		 \\	
\midrule
\multirow{10}{*}{Generic} 
& Tracking Shuffled Objects [three] & 0.027&	0.993 & 0.510&	0.020	& 0.980 \\
& Disambiguation QA & 0.397&	0.991&	0.694 & 0.388&	0.612	\\
& Web of Lies &0.187 &	0.867&	0.527 & 0.053&	0.947	\\
& Temporal Sequences & 0.150&	0.983& 0.566 & 0.133&	0.867	\\
& Sports Understanding & 0.394&	0.830&	0.612 & 0.223&	0.777	\\
& Salient Translation Error Detection & 0.490&	0.867&	0.679& 0.357&	0.643	\\
& Penguins in a Table &  0.527&	 0.893&	 0.710 & 0.420 &	0.580 	\\
& Logical Deduction [three] & 0.069 &	0.983 &	0.526 & 0.052 &	0.948 	\\
& Navigate & 0.226 &	0.884 &	0.555  & 0.110 &	0.890 	\\
\cmidrule{2-7}
& Avg. & 0.274 &	0.921 &	0.598 & 0.195 &	0.805 	 \\	
\bottomrule
\end{tabular}
}
\caption{Evaluation Results for Each of the Benchmarks in the LLaMA3 Model}
\label{table:llama3}
\end{table*}

\begin{table*}[!htbp]
\centering
\small
\resizebox{\linewidth}{!}
{
\begin{tabular}{clccccc}
\toprule
\textbf{Category} & \textbf{Benchmark} & \textbf{LLM Correct} & \textbf{LLM False} & \textbf{Average} & \textbf{Both} & \textbf{Either} \\
\toprule
Mathematics & GSM8K &  1.000 &	0.127 & 0.564 & 0.127 &	0.873 	  \\
\midrule
First-Order Logic & PrOntoQA & 0.459 &	0.433 & 0.446 	& 0.215 & 0.463 \\
\midrule
\multirow{4}{*}{Commonsense} 
& StrategyQA & 1.000 & 0.219& 0.609  &	0.219 &	0.781 	 \\
& CommonsenseQA 2.0 & 1.000 & 0.138&	0.569 & 0.138&	0.862 \\
& Creak & 1.000 &	0.192 & 0.569 & 0.138 & 0.862 \\
\cmidrule{2-7}
& Avg. & 1.000 &	0.183 &	0.592& 0.183 & 0.817		 \\	
\midrule
\multirow{10}{*}{Generic} 
& Tracking Shuffled Objects [three] & 1.000&	0.108& 0.554 & 0.108&	0.892	 \\
& Disambiguation QA & 0.991&	0.240&	0.616 & 0.207&	0.784	\\
& Web of Lies &1.000 &	0.207&	 0.603& 0.207&0.784		\\
& Temporal Sequences & 1.000&	0.098& 0.549 & 0.098&0.902	\\
& Sports Understanding & 1.000&	0.223&	0.611 & 0.218&	0.782	\\
& Salient Translation Error Detection & 1.000&	0.286&	0.643& 0.286&0.714		\\
& Penguins in a Table &  1.000&0.301	 &0.650	  & 0.277 &0.723	 	\\
& Logical Deduction [three] & 1.000 &	0.110 &	0.555 & 0.110 &	0.890 	\\
& Navigate & 1.000 &	0.103 &	0.551  & 0.103 &	0.897 	\\
\cmidrule{2-7}
& Avg. & 0.999 &	0.186 &0.593	 &0.179  &0.820	 	 \\	
\bottomrule
\end{tabular}
}
\caption{Evaluation Results for Each of the Benchmarks Using Verbalization in the LLaMA3 Model}
\label{table:llama3+verbaliztion}
\end{table*}

\begin{table*}[!htbp]
\centering
\small
\resizebox{\linewidth}{!}
{
\begin{tabular}{clccccc}
\toprule
\textbf{Category} & \textbf{Benchmark} & \textbf{LLM Correct} & \textbf{LLM False} & \textbf{Average} & \textbf{Both} & \textbf{Either} \\
\toprule
Mathematics & GSM8K &  0.216 &	1.000 & 0.608& 0.216 &	0.784 	  \\
\midrule
First-Order Logic & PrOntoQA & 0.700 &	0.765 &  0.733	& 0.466 & 0.534 \\
\midrule
\multirow{4}{*}{Commonsense} 
& StrategyQA & 0.405 & 0.921 & 0.663  &	0.326 &	 0.674	 \\
& CommonsenseQA 2.0 & 0.396 & 0.915&	0.663 & 0.326&	 0.674\\
& Creak & 0.248 &	0.954 & 0.601 & 0.201 & 0.799 \\
\cmidrule{2-7}
& Avg. & 0.349 &	0.930 &	0.640& 0.279 &	0.721	 \\	
\midrule
\multirow{10}{*}{Generic} 
& Tracking Shuffled Objects [three] & 0.142&	1.000& 0.571 & 0.142&	0.858	 \\
& Disambiguation QA & 0.750&	0.710&	0.730 & 0.460&	0.540	\\
& Web of Lies &0.833	&	0.727 & 0.780&	0.560&0.440	\\
& Temporal Sequences &0.486 &	1.000& 0.743 & 0.486&	0.514	\\
& Sports Understanding & 0.614&	0.935&	0.774 & 0.549&	0.451	\\
& Salient Translation Error Detection &0.418 &	0.990&	0.704& 0.408&0.592		\\
& Penguins in a Table &  0.272&	1.000 &0.636	  &0.272  &	 0.728	\\
& Logical Deduction [three] & 0.424 &	0.948 &	0.686 & 0.372 &	0.628 	\\
& Navigate & 0.274 &	0.993 &	0.634  & 0.267 &	0.733 	\\
\cmidrule{2-7}
& Avg. & 0.468 &	0.922 &	0.695 & 0.391 &	0.609 	 \\	
\bottomrule
\end{tabular}
}
\caption{Evaluation Results for Each of the Benchmarks Using Semantic Entropy in the LLaMA3 Model}
\label{table:llama3+se}
\end{table*}

\begin{table*}[!htbp]
\centering
\small
\resizebox{\linewidth}{!}
{
\begin{tabular}{clccccc}
\toprule
\textbf{Category} & \textbf{Benchmark} & \textbf{LLM Correct} & \textbf{LLM False} & \textbf{Average} & \textbf{Both} & \textbf{Either} \\
\toprule
Mathematics & GSM8K &  0.220 &	0.981 & 0.600& 0.200 &	0.800 	  \\
\midrule
First-Order Logic & PrOntoQA & 0.036 &	1.000 & 0.518 	& 0.036 & 0.964 \\
\midrule
\multirow{4}{*}{Commonsense} 
& StrategyQA &0.502  &0.603 & 0.563  &	0.130 &	0.865 	 \\
& CommonsenseQA 2.0 & 0.415 & 0.692&	0.554 & 0.112&	0.885 \\
& Creak & 0.449 &	0.656 & 0.553 & 0.108 & 0.889 \\
\cmidrule{2-7}
& Avg. & 0.456 &	0.657 &	0.556& 0.117 &0.879		 \\	
\midrule
\multirow{10}{*}{Generic} 
& Tracking Shuffled Objects [three] & 0.115&	1.000& 0.557 & 0.115&	0.885	 \\
& Disambiguation QA & 0.448&	1.000&0.724	 &0.448 &0.552		\\
& Web of Lies & 0.220&	0.913&	 0.567& 0.133&0.867		\\
& Temporal Sequences & 0.220&	1.000& 0.610 &0.220 &	0.780	\\
& Sports Understanding & 0.420&		0.862 & 0.641&	0.282	& 0.718\\
& Salient Translation Error Detection & 0.602&	0.969&	0.786& 0.571&	0.429	\\
& Penguins in a Table &  0.643&	1.000 &	0.821  & 0.643 &0.357	 	\\
& Logical Deduction [three] &  0.139&	1.000 &	0.569 & 0.139 &	0.861 	\\
& Navigate &  0.301&	0.973 &	0.637 & 0.274 &	0.726 	\\
\cmidrule{2-7}
& Avg. &  0.345&	0.969 &	0.657  & 0.314 &	0.686 	\\
\bottomrule
\end{tabular}
}
\caption{Evaluation Results for Each of the Benchmarks Using P(True) in the LLaMA3 Model}
\label{table:llama3+p_true}
\end{table*}

\begin{table*}[!htbp]
\centering
\small
\resizebox{\linewidth}{!}
{
\begin{tabular}{clccccc}
\toprule
\textbf{Category} & \textbf{Benchmark} & \textbf{LLM Correct} & \textbf{LLM False} & \textbf{Average} & \textbf{Both} & \textbf{Either} \\
\toprule
Mathematics & GSM8K & 0.287  &	1.000 & 0.643& 0.287 &	0.713 	  \\
\midrule
First-Order Logic & PrOntoQA & 0.147 &	1.000 & 0.573 	& 0.147 & 0.853 \\
\midrule
\multirow{4}{*}{Commonsense} 
& StrategyQA & 0.628 & 0.753& 0.691  &	0.386 &	0.609 	 \\
& CommonsenseQA 2.0 & 0.515 & 0.796&	0.656 & 0.315&	0.681 \\
& Creak & 0.536 &	0.734 & 0.635 & 0.272 & 0.724 \\
\cmidrule{2-7}
& Avg. & 0.560 &	0.761 &	0.660& 0.325 &	0.672	 \\	
\midrule
\multirow{10}{*}{Generic} 
& Tracking Shuffled Objects [three] & 0.297&	1.000& 0.649 &0.297 &	0.703	 \\
& Disambiguation QA & 0.534&	1.000&	0.767 &0.534 &	0.466	\\
& Web of Lies &0.287 &	0.967&	 0.627& 0.253&	0.747	\\
& Temporal Sequences & 0.382&	1.000& 0.691 & 0.382&		0.618\\
& Sports Understanding & 0.473&	0.915&	0.694 & 0.388&	0.612	\\
& Salient Translation Error Detection & 0.561&	0.918&0.740	& 0.480&	0.520	\\
& Penguins in a Table & 0.580 &	0.946 &	 0.763 & 0.527 &	 0.473	\\
& Logical Deduction [three] & 0.301 &	1.000 &	0.650 & 0.301 &0.699	 	\\
& Navigate &  0.336&	0.993 &	0.664  &0.329  &0.671	 	\\
\cmidrule{2-7}
& Avg. & 0.417 &	0.971 &	0.694 &0.388  &	0.612 	 \\	
\bottomrule
\end{tabular}
}
\caption{Evaluation Results for Each of the Benchmarks Using Predictive Entropy in the LLaMA3 Model}
\label{table:llama3+predictive entropy}
\end{table*}

\begin{table*}[!htbp]
\centering
\small
\resizebox{\linewidth}{!}
{
\begin{tabular}{clccccc}
\toprule
\textbf{Category} & \textbf{Benchmark} & \textbf{LLM Correct} & \textbf{LLM False} & \textbf{Average} & \textbf{Both} & \textbf{Either} \\
\toprule
Mathematics & GSM8K & 0.304  &	1.000 & 0.652& 0.304 &	0.696 	  \\
\midrule
First-Order Logic & PrOntoQA &  0.723&	0.782 &  0.752	& 0.505 & 0.495 \\
\midrule
\multirow{4}{*}{Commonsense} 
& StrategyQA & 0.698 & 0.809 & 0.753  &	0.512 &	0.484 	 \\
& CommonsenseQA 2.0 & 0.569 & 0.838 &	0.704 & 0.412&	 0.585\\
& Creak & 0.567 &	0.771 & 0.669 & 0.341 & 0.656 \\
\cmidrule{2-7}
& Avg. & 0.611 &	0.806 &	0.709 & 0.421 &	0.575	 \\	
\midrule
\multirow{10}{*}{Generic} 
& Tracking Shuffled Objects [three] & 0.365&	1.000& 0.682 & 0.365&	0.635	 \\
& Disambiguation QA & 0.578&	1.000&0.789	 &0.578 &	0.422	\\
& Web of Lies & 0.313&	1.000&0.657	 &0.313 &0.687		\\
& Temporal Sequences & 0.491&	1.000& 0.746 & 0.491&	0.509	\\
& Sports Understanding & 0.647&	0.973&	0.810 & 0.620&0.380		\\
& Salient Translation Error Detection &0.622 &1.000	&	0.811& 0.622&0.378		\\
& Penguins in a Table &  0.625&	0.982 &	0.804  & 0.607 &	0.393 	\\
& Logical Deduction [three] & 0.442 &	0.953 &	0.698 & 0.395 &	0.605 	\\
& Navigate & 0.356 &	 1.000&	 0.678 & 0.356 &	0.644 	\\
\cmidrule{2-7}
& Avg. & 0.493 &	0.990 &	0.742 & 0.483 &	0.517 	 \\	
\bottomrule
\end{tabular}
}
\caption{Evaluation Results for Each of the Benchmarks Using AFICE in the LLaMA3 Model}
\label{table:llama3+ AFICE}
\end{table*}

\end{document}